\RequirePackage{fix-cm}

\documentclass[twocolumn]{svjour3}          % twocolumn
\smartqed  % flush right qed marks, e.g. at end of proof
\usepackage{graphicx}
\usepackage[bookmarks=true, colorlinks, linkcolor=red, anchorcolor=blue, citecolor=green]{hyperref}
\usepackage{times}
\usepackage{amsmath}
\usepackage{amssymb}
\usepackage{commath}
\usepackage{tabularx}
\usepackage{booktabs}
\usepackage{float}
\usepackage{algorithm}
\usepackage{algpseudocode}
\usepackage{amsmath}
\usepackage{graphicx}
\usepackage{subfigure}
\usepackage{color}
\usepackage{CJK}
\usepackage{cite}
\usepackage{multirow}

\usepackage{blindtext}
\begin{document}
\begin{sloppypar}
\title{Iterative Clustering with Game-Theoretic Matching for Robust Multi-consistency Correspondence}

%\titlerunning{Short form of title}        % if too long for running head

\author{Chen Zhao        \and
        Jiaqi Yang \and
        Ke Xian \and
        Zhiguo Cao \and
        Xin Li
}

%\authorrunning{Short form of author list} % if too long for running head

\institute{Chen Zhao\and Jiaqi Yang\and Ke Xian\and Zhiguo \at
              School of Artificial Intelligence and Automation \\
              Huazhong University of Science and Technology \\
              Wuhan, 430074, China \\
              \email{hust\_zhao@hust.edu.cn, jqyang@hust.edu.cn, \\ kexian@hust.edu.cn, zgcao@hust.edu.cn}           %  \\
%             \emph{Present address:} of F. Author  %  if needed
           \and
           Xin Li \at
              Electrical Engineering \\ 
              West Virginia University \\
              Morgantown, 26506, USA \\
              \email{Xin.Li@mail.wvu.edu}
}

\date{Received: date / Accepted: date}
% The correct dates will be entered by the editor

\maketitle

\begin{abstract}
Matching corresponding features between two images is a fundamental task to computer vision with numerous applications in object recognition, robotics, and 3D reconstruction. Current state of the art in image feature matching has focused on establishing a single consistency in static scenes; by contrast, finding multiple consistencies in dynamic scenes has been under-researched. In this paper, we present an end-to-end optimization framework named ``iterative clustering with Game-Theoretic Matching'' (ic-GTM) for robust multi-consistency correspondence. The key idea is to formulate multi-consistency matching as a generalized clustering problem for an image pair. In our formulation, several local matching games are simultaneously carried out in different corresponding block pairs under the guidance of a novel payoff function consisting of both geometric and descriptive compatibility; the global matching results are further iteratively refined by clustering and thresholding with respect to a payoff matrix. We also propose three new metrics for evaluating the performance of multi-consistency image feature matching. Extensive experimental results have shown that the proposed framework significantly outperforms previous state-of-the-art approaches on both single-consistency and multi-consistency datasets.
\keywords{Image feature matching \and Multiple consistencies \and Robust correspondence \and Game theory}
% \PACS{PACS code1 \and PACS code2 \and more}
% \subclass{MSC code1 \and MSC code2 \and more}
\end{abstract}

\section{Introduction}
\label{sec:intro}
Image feature matching (a.k.a. correspondence selection) is a cornerstone of many computer vision and robotic tasks, such as optical flow~\cite{baker2011database}, structure-from-motion~\cite{snavely2008modeling}, stereo matching~\cite{hirschmuller2008stereo}, simultaneous localization and mapping~\cite{benhimane2004real}, and image stitching~\cite{brown2007automatic}. The main purpose of image feature matching is to discover the corresponding relationship between feature points of two images, which serves as the foundation for analysis at higher levels. Despite being a trivial task for human vision, image feature matching becomes more challenging for machines especially in the presence of large variation in illumination and viewpoint. It is often non-trivial to pursue robust features invariant to illumination and viewpoint. 
\begin{figure}[t]
	\centering
	\subfigure[Single-consistency]
	{ \includegraphics[width=0.47\linewidth]{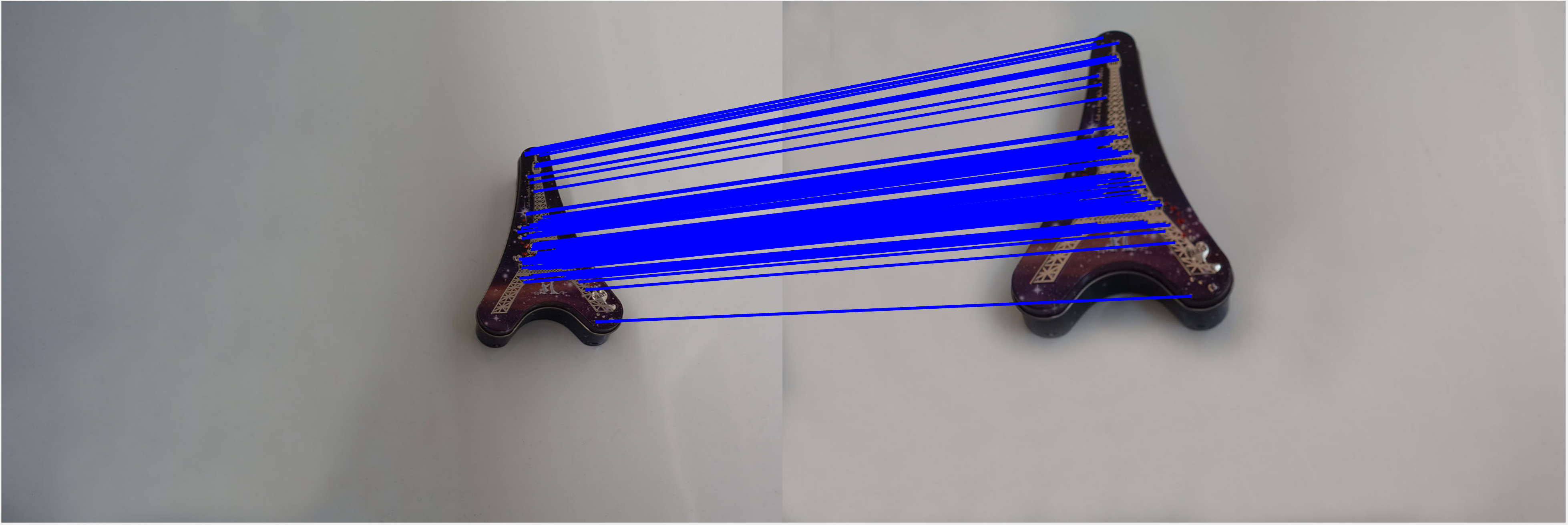}\label{fig_secondsub} }
	\subfigure[Multi-consistency]
	{ \includegraphics[width=0.47\linewidth]{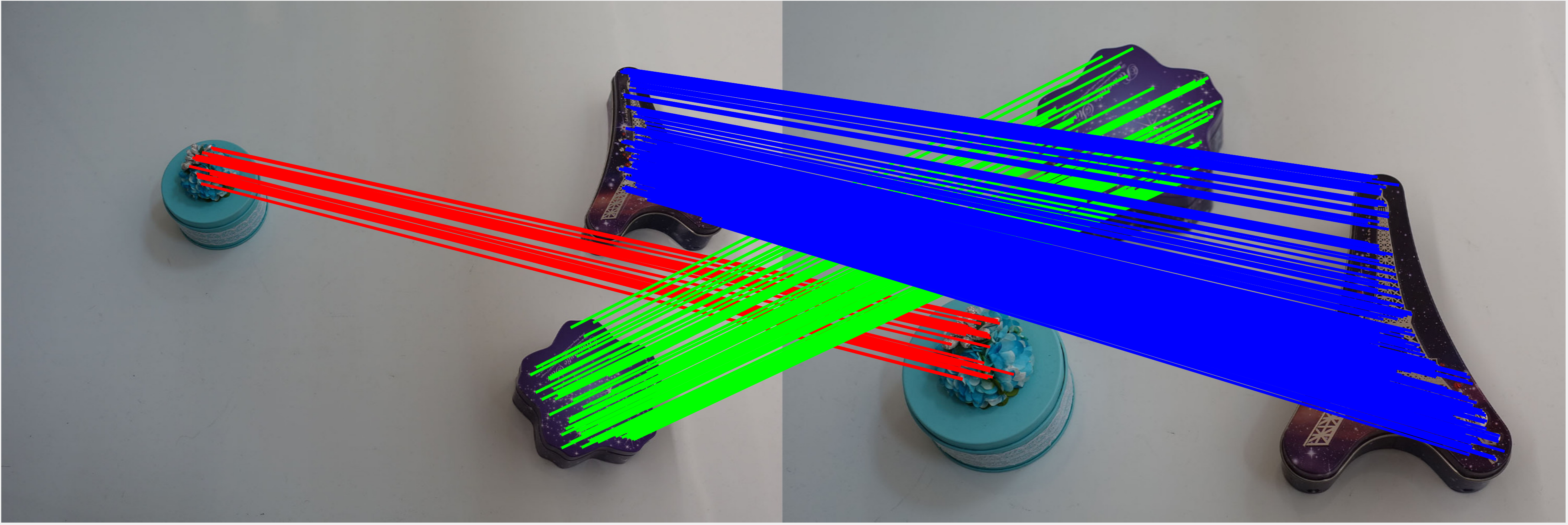}\label{fig_thirdsub} }
	\caption{Illustrations of single-consistency feature matching and multi-consistency feature matching.}
	\label{fig:fig1}
\end{figure}
The problem is further complicated when the scene is dynamic instead of static. The majority of existing approaches have been developed for static scenes only - i.e., the corresponding relationship between two images is characterized by a global transformation (e.g., affine transformation and perspective transformation, as shown in Fig.~\ref{fig:fig1}a). Such single-consistency feature matching is not appropriate for dynamic scenes in which there are multiple separated local transformations associated with several moving objects (e.g., Fig.~\ref{fig:fig1}b). Note that due to the existence of multiple consistency, conventional wisdom of improving the robustness of single-consistency matching such as RANSAC ~\cite{fischler1981random} and USAC (the modified version of RANSAC)~\cite{Raguram2013USAC} easily fails. %As shown in Sec.~\ref{sec:exper}, RANSAC and its modified algorithms are prone to missing some less salient consistencies. 

In this project, we approach the problem of multi-consistency image feature matching by formulating it as a generalized clustering problem. The key insight behind our approach lies in that multiple consistencies between two images are determined by a collection of homographies corresponding to either planar surface in the background or independent moving objects in the foreground in dynamic scenes. In addition to correspondence establishment, determining the total number of consistencies homographies $K$ is a new issue that has not been addressed in the open literature. By taking error-free correspondences between two images as input features, one can solve the problem by clustering in the feature space in a similar fashion to k-means (note that the number of clusters $k$ is often specified by the user). In view of practical limitations (i.e., local correspondences are error-prone), we have to develop robust clustering solutions insensitive to the possible outliers in local correspondences.  

The motivation behind our approach is two-fold. On one hand, game-theoretic matching (GTM)~\cite{albarelli2010game} has been developed as a powerful technique for establishing single-consistency correspondence even in the presence of elastic deformation ~\cite{Rodola2014Elastic}; however, it has not been extended to multiple consistencies to the best of our knowledge. The conventional framework of GTM is inappropriate for multi-consistency feature matching because the inliers associated with different moving objects are incompatible, which violates the fundamental assumption with a global geometric compatibility between feature correspondences. To overcome this limitation, we propose a novel payoff function that considers both \emph{geometric} and \emph{descriptive} compatibility in the definition of payoff function. With newly defined payoff function, we can play multiple local games simultaneously by following the classical evolutionary stable strategy (ESS) algorithm \cite{weibull1997evolutionary}.

%Actually, there have been some non-parametric approaches that enable to play the same role as multiple local games and are insensitive to the multi-consistency challenge. Nerveless, these methods, such as VFC~\cite{Ma2014Robust}, GMS~\cite{bian2017gms}, and LPM~\cite{ma2019locality}, typically rely on some prior assumptions, -e.g., Gaussian Distribution~\cite{Ma2014Robust} and motion smoothness~\cite{bian2017gms}, which can be invalid and result in a limited tolerance to large appearance changes. Moreover, the putative correct correspondences selected by these non-parametric algorithms are regarded as a single consistency, with the potential multiple consistencies being not recognized. 

On the other hand, we propose an iterative consistency clustering procedure to group compatible correspondences and estimate the unknown number of clusters $K$ based on the results of local non-cooperative games. Since the compatibility between each two tentative matches determined by local games is measured by the newly defined payoff function, it appears plausible to use this compatibility-based metric to group those correspondences with high compatibility and infer the local transformation generated from each correspondence cluster. Conceptually similar to k-means, we can alternate the steps of compatibility-based clustering and local transformation estimation. During the iterations, image feature pairs falsely eliminated in local games can be recovered by clustering. More specifically, by calculating the consistency with estimated local transformations, we can recover inliers because they should be consistent with at least one estimated local transformation. Such iteration can be terminated whenever no new transformation can be found (i.e., the estimated $K$ reaches the maximum). %The ablation study in Sec.~\ref{sec:exper} has been indicated that the number of true positive correspondences surviving after the selection significantly increases with this procedure being added. 

The other contribution of this work is on performance evaluation for multi-consistency feature matching. In conventional single-consistency matching, three metrics, -i.e., precision (P), recall (R), and F-measure (F), have been used in~\cite{lin2014bilateral, bian2017gms, ma2019locality}; but they are sensitive to the unbalanced saliency of different underlying consistencies and therefore inappropriate for evaluating the performance of multi-consistency feature matching. Note that the distribution of keypoint-level correspondences are often sparse and nonuniform; therefore, the number of included correspondences is likely significantly vary from cluster to cluster. To address this issue, we propose to use three new metrics for multi-consistency evaluation - i.e., weighted-precision (W-P), weighted-recall (W-R), and weighted-F-measure (W-F). The key idea is to adaptively weight each correspondence based on the underlying consistency, with the aim of amplifying the effect of less salient consistencies. The implementation details of the benchmark\footnote{The code and dataset will be available at: \url{https://github.com/sailor-z/ic-GTM}.} will be introduced in Sec.~\ref{sec:exper}. \footnote{This paper is an extended version of the conference paper~\cite{zhao2018scalable}}.

In a nutshell, the contributions of this paper are summarized as follows:

\begin{itemize}
	\item A formulation of multi-consistency image feature matching problem and theoretical analysis about its relationship to single-consistency matching and generalized clustering.
	\item A novel payoff function robust to common disturbance to guide both playing local games and clustering global consistencies, with the consideration of both geometric and descriptive compatibility.
	\item An iterative clustering with Graph-Theoretic Matching (ic-GTM) framework for multi-consistency image feature matching, which has significantly outperformed other competing methods on both the multi-consistency and single-consistency datasets.
\end{itemize} 

\section{Related work}
\label{sec:relwork}
%In this section, we briefly review some correspondence selection algorithms that can be divided into three categories as follows. 
\textbf{Parametric algorithms.  } A popular strategy for correspondence selection is based on the classical RANSAC~\cite{Raguram2013USAC}. In RANSAC, one alternately samples a subset of correspondences to generate a hypothesized parametric model and verifies the confidence of the generated model by some geometric metrics (e.g., reprojected errors and epipolar distances). These metrics can also be used to select consistent correspondences under the constraint of the finally estimated model. However, the hypothesis generated by random sampling is sensitive to the inlier ratio of initial correspondence set. The confidence of hypothesis testing tends to decline rapidly if the majority of initial correspondences are incorrect. Although some efforts -e.g., PROSAC~\cite{chum2005matching}, LO-RANSAC~\cite{chum2003locally}, and USAC~\cite{Raguram2013USAC}, have been proposed to improve the robustness to low initial inlier ratios, parametric methods still have fundamental limitations in the scenarios of non-rigid feature matching and multi-consistency feature matching. 

For non-rigid feature matching, the underlying transformation between two images is too complex to be accurately represented by a global transformation, -e.g., homography matrix or essential matrix. For multi-consistency feature matching, some researches fit multiple parametric models by generating multiple hypothesis sampling~\cite{magri2015robust, chin2010accelerated, wong2011dynamic}. These approaches generally suppose that correspondences associated with the same structure share a common hypotheses. For example, Chin et al. proposed a guided-sampling scheme (Multi-GS)~\cite{chin2010accelerated} where a series of hypotheses are generated in advance and the preference lists ordered according to the compatibility to the hypotheses are expected to be similar. However, image feature matching and image segmentation are inter-twisted in multi-consistency (i.e., like a chicken-and-egg problem).
\\
\\
\textbf{Non-parametric algorithms.  } An alternative approach to correspondence selection is via non-parametric models~\cite{albarelli2010game, Ma2014Robust, bian2017gms, ma2019locality}. For instance,  in game-theoretic matching (GTM)~\cite{albarelli2010game},  inliers are assumed to be compatible with each other, which result in a larger payoff in a non-cooperative game. The vector filed consensus (VFC)~\cite{Ma2014Robust} approach is based on the assumption that noise around inliers and outliers observes the Gaussian distribution and the uniform distribution respectively. It follows that a maximum a posteriori (MAP) estimation of a mixture model with latent variables determining inliers is obtained by the EM algorithm. Grid-based motion statistics (GMS)~\cite{bian2017gms} is based on the observation that the matching quality is positively correlated to the number of correspondences in small grid regions under the assumption with motion smoothness. Most recently, a locality preserving matching  (LPM)~\cite{ma2019locality} method was developed based on the observation that the spatial neighborhood relationship between two keypoints of a correct matching should be well preserved. Although these non-parametric approaches can be applied to multi-consistency feature matching, their performance is prone to declining in the context of some specific challenges, -e.g., the large scale rotation, translation, or/and zoom. %Moreover, although the putative inliers can be selected by these algorithms, the underlying consistencies represented by some parametric transformations among the inliers are not recognized. 
It remains an open research problem how to discover and recognize the potential consistent relationship from the selected correspondences in nonparametric methods.   
\\
\\
\textbf{Learning-based algorithms.  } Recently, deep learning has found many successful applications in image processing and computer vision such as image classification, image segmentation, object detection and recognition. Naturally, it is desirable to pursue a learning-based approach toward correspondence selection. Some attempts~\cite{yi2018learning, zhao2019nm} have been made along this direction, which translated the correspondence selection problem to a per-match binary classification problem (i.e., inlier vs. outlier). However, these supervised leaning-based approaches require enormous annotated training data; acquiring such annotations for the multi-consistency feature matching task is often impractical because the workers have to manually label thousands of correspondences in an image pair. %Therefore, in the absence of sufficient labeled training data, it is highly non-trivial to train a powerful deep network to search for multiple consistencies. 

\section{Method}
\begin{figure}[t]
	\centering
	\includegraphics[width=1.0\linewidth]{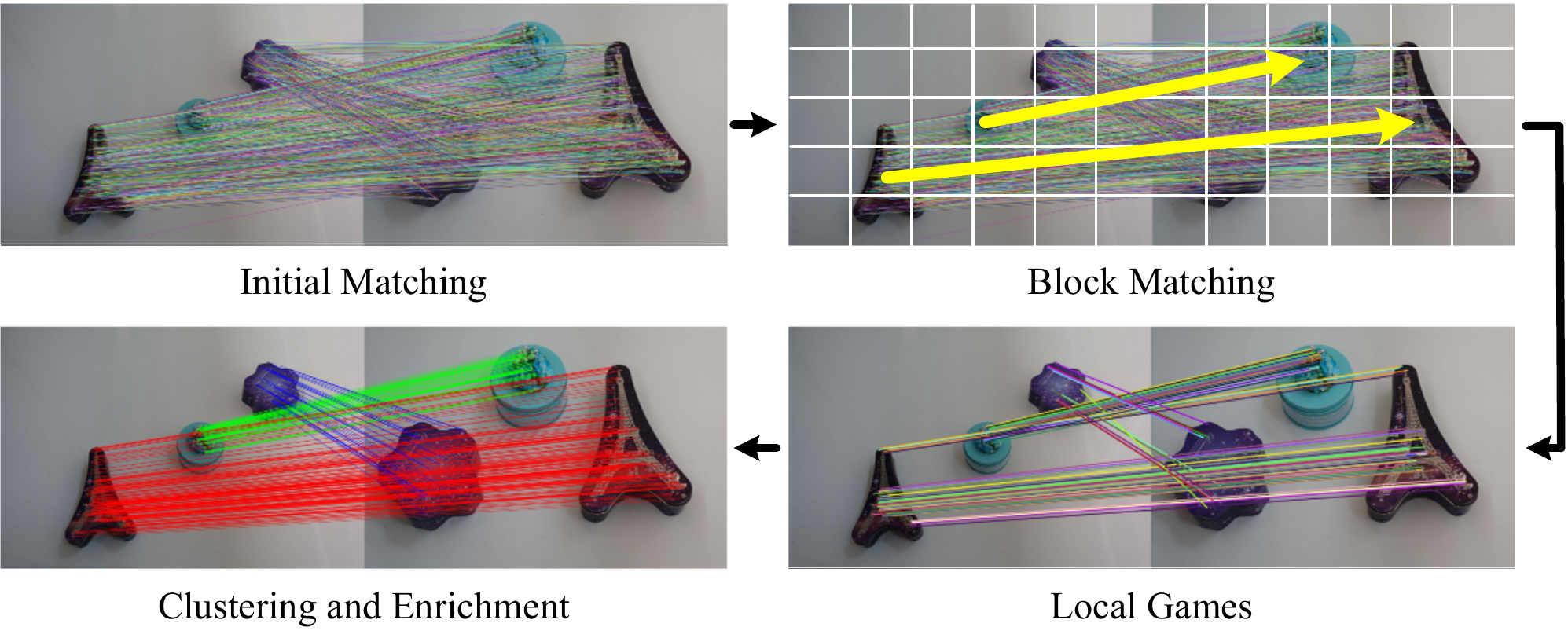}
	\caption{Overview of the proposed framework for multi-consistency feature matching -i.e., Initialization, Block Matching, Local Games, and Iterative Clustering. Yellow arrows connect the grouped block pairs during block matching and lines with different colors represent the final result of multiple consistencies discovered by ic-GTM.}
	\label{fig:way}
\end{figure}
Fig.~\ref{fig:way} includes the overview of our ic-GTM framework consisting of four steps. Initialization step generates initial correspondences prone to a large number of mismatches; The step of Block Matching (Sec.~\ref{subsec:grid}) divides the images into non-overlapping blocks and searches for the block pairs; Local Game step (Sec.~\ref{subsec:game-theory}) carries out a series of non-cooperative games with all block pairs simultaneously and identifies plausible candidates; finally, Iterative Clustering  (Sec.~\ref{subsec:enrich}) clusters the correspondences survived from local games and recovers the incorrectly discarded inliers in an iterative manner.

\subsection{Problem Formulation}
Given a pair of images $(I, I^{'})$, detected keypoints $(\mathcal{K}=\{\mathbf{k}_i\}, \mathcal{K}^{'}=\{\mathbf{k}^{'}_i\})$, and local patch descriptions $(\mathcal{F}=\{\mathbf{f}_i\}, \mathcal{F}^{'}=\{\mathbf{f}^{'}_i\})$ $(i=1, 2, ..., N)$, an initial correspondence set $\mathcal{C}=\{c_i, c_i=\{(\mathbf{k}_i, \mathbf{k}^{'}_i), (\mathbf{f}_i, \mathbf{f}_i^{'})\}\}$ can be generated by some ad-hoc strategies such as brute force matching between $\mathcal{F}$ and $\mathcal{F}^{'}$. In the situation of dynamic scenes, $\mathcal{C}$ tends to contain multiple error-prone consistencies representing different moving objects in the foreground. Due to inevitable errors of keypoint detection and intrinsic ambiguity of feature descriptions, there are many nuisances in $\mathcal{C}$, -i.e., outliers $\mathcal{C}_{outlier}$. The goal of multi-consistency matching is to reject $\mathcal{C}_{outlier}$  while identifying the correct subset of $\mathcal{C}$ (i.e., $\mathcal{C}_{inlier}$). Then it will be straightforward to estimate a set of local transformations $\mathcal{H}={h_i}$ $(i=1, 2, ..., K)$ based on the multi-consistency matching result. We will elaborate the details of three steps in the following subsections.
\subsection{Block Matching}
\label{subsec:grid}
Directly applying RANSAC~\cite{fischler1981random} or game-theoretic matching (GTM)~\cite{albarelli2010game} is ill-suited for multi-consistency feature matching due to the lack of a single global transformation. Based on this observation, it is natural to work with local regions instead of the image as a whole. One possible solution is to leverage off-the-shelf image segmentation algorithm, but we note that  segmentation is an over-kill for multi-consistency matching. As a compromised solution, we propose a simple yet effective gridding method that divides the image into non-overlapping blocks ($\mathcal{G}=\{g_i\}, i=1, 2, ..., M$). Then we can search for the corresponding block pairs $(g_i, g_i^{'})$ between two images, with the expectation that each pair of matched blocks only contains a single consistency. 

In the presence of large-scale rigid transformation between $(I, I^{'})$ (e.g., viewpoint changes and rotation), it is non-trivial to exactly search for the corresponding block $g_i^{'}$ in $I^{'}$ for each $g_i$ in $I$. Drawing inspiration from~\cite{bian2017gms} that suggests the confidence of a correspondence is locally correlated with the number of matches, we propose to address this issue in a statistical manner. That is, the similarity of each block pair is quantified by the number of correspondences located within the pair. Formally, we have
\begin{equation}\label{eq:LRF1}
{S(g_i,g_i^{'})}=\sum\limits_{j}{p_j},
\end{equation}
where $S(g_i,g_i^{'})$ is the similarity of blocks $(g_i,g_i^{'})$ and correspondence $p_j$ is defined by
\begin{equation}\label{eq:LRF2}
{{p}_{j}}=\left\{ \begin{aligned}
& 1, \text{ if } {{c}_{j}} \text{ is located in } (g_i,g_i^{'}) \\ 
& 0, \text{ otherwise} \\ 
\end{aligned} \right.,
\end{equation}
with $c_j \in {\mathcal{C}}$. For each block $g_a$ in $I$, the corresponding block $g_b$ in $I^{'}$ is given by the most similar one or the one with the largest number of correspondences - i.e.,
\begin{equation}\label{eq:LRF5}
    g_b = \underset{g_b}{arg max}\ S(g_a, g_b).
\end{equation}

Meantime, if the number of correspondences contained in the matched block pair is smaller than a pre-defined threshold, the pair is discarded because it is likely caused by the interference from background or clutter. For example, the grouping tends to be ambiguous in some blocks such as those including the edge or corner of an object as shown in the top right image of Fig.~\ref{fig:way}. The matched blocks found in these regions are prone to be eliminated due to a small number of correspondences. 
%To remedy this issue, we add an enrichment step to retrieve the missed correspondences, which will be introduced in Sec.~\ref{subsec:enrich}.

\subsection{Play Local Games}
\label{subsec:game-theory}
The step of block matching supplies multiple consistencies assigned to different regions. However, those tentative matching results have not been optimized. Inspired by the success of GTM for single-consistency matching, it is intuitively desirable to optimize the matching results by playing several local games simultaneously in these local regions. 
Similar to the GTM for single-consistency matching~\cite{albarelli2010game}, at the core of each local game is the payoff function which represents the compatibility of two correspondences. Players who achieve higher payoffs are more popular as the game evolves, which suggests the correspondences they select are inliers. However, the payoff function employed in traditional GTM only considers the geometric compatibility between two correspondences, whose reliability becomes questionable when large transformation is present. To overcome this limitation, we propose a more robust payoff function considering both geometric and descriptive compatibility in this work. 

More specifically, each player chooses a correspondence $c=\{(\mathbf{k},\mathbf{k^{'}}),(\mathbf{f},\mathbf{f}^{'})\}$ from $\mathcal{C}$, where $(\mathbf{k},\mathbf{k^{'}})$ denotes the pair of matched keypoints and $(\mathbf{f},\mathbf{f}^{'})$ is the corresponding pair of local descriptions. Each two players will then receive a payoff function positively correlated with the compatibility of their choices. The payoff function is defined by
%\begin{equation}\label{eq:LRF3}
%{{P}_{ij}}=P_{ij}^{geo}+P_{ij}^{des}+\lambda \left| P_{ij}^{geo}-P_{ij}^{des} \right|,
%\end{equation}
\begin{equation}\label{eq:LRF3}
    {{P}_{ij}}=P_{ij}^{geo}+P_{ij}^{des},
\end{equation}
where $P_{ij}$ is the overall payoff of $(c_i,c_j)$, $P_{ij}^{geo}$ and $P_{ij}^{des}$ respectively indicate the compatibility of geometric structures and local descriptions, which we will elaborate next. 
\\
\\
\begin{figure}[t]
	\centering
	\subfigure[Inliers]
	{ \includegraphics[width=0.47\linewidth]{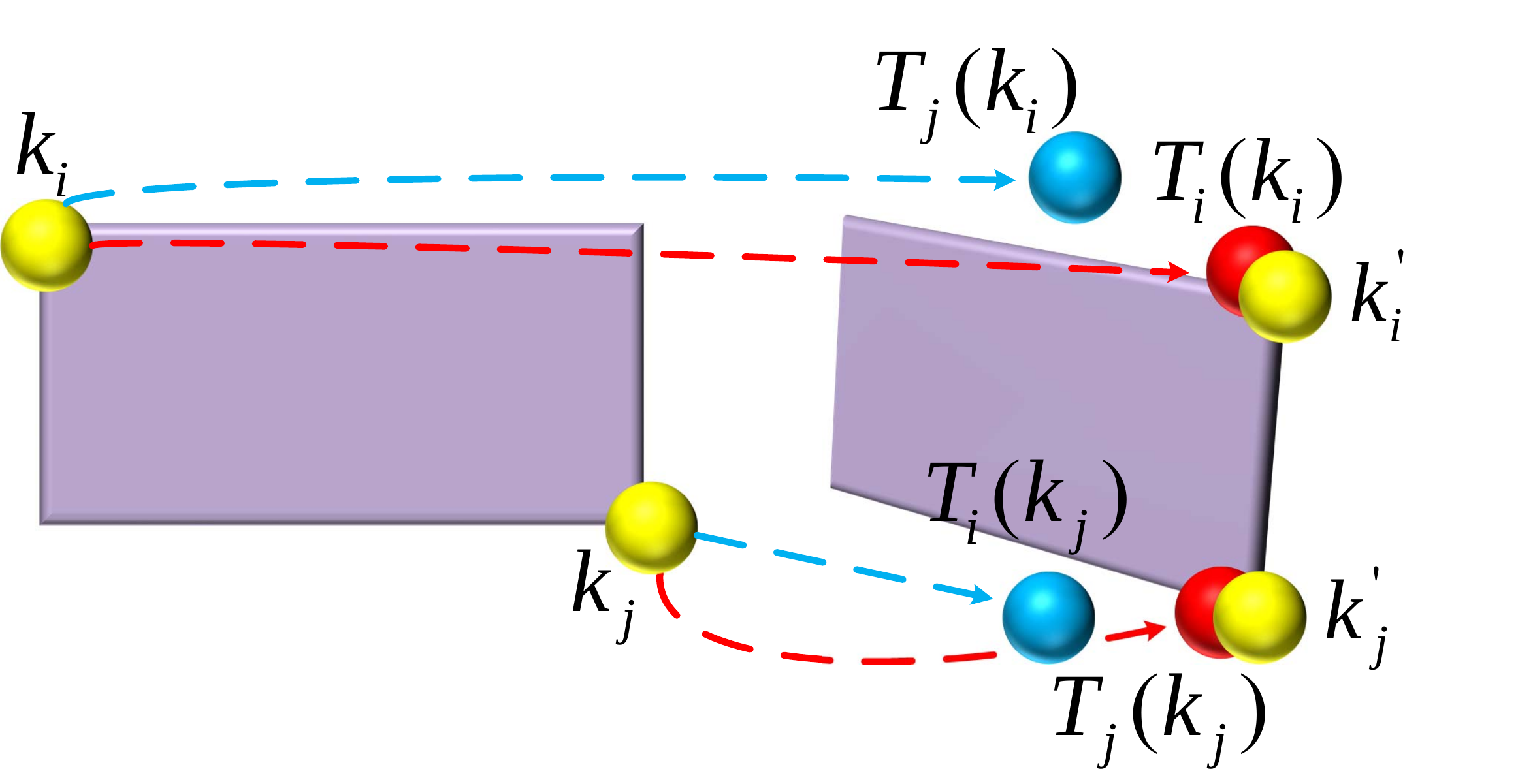} }
	\subfigure[Outliers]
	{ \includegraphics[width=0.47\linewidth]{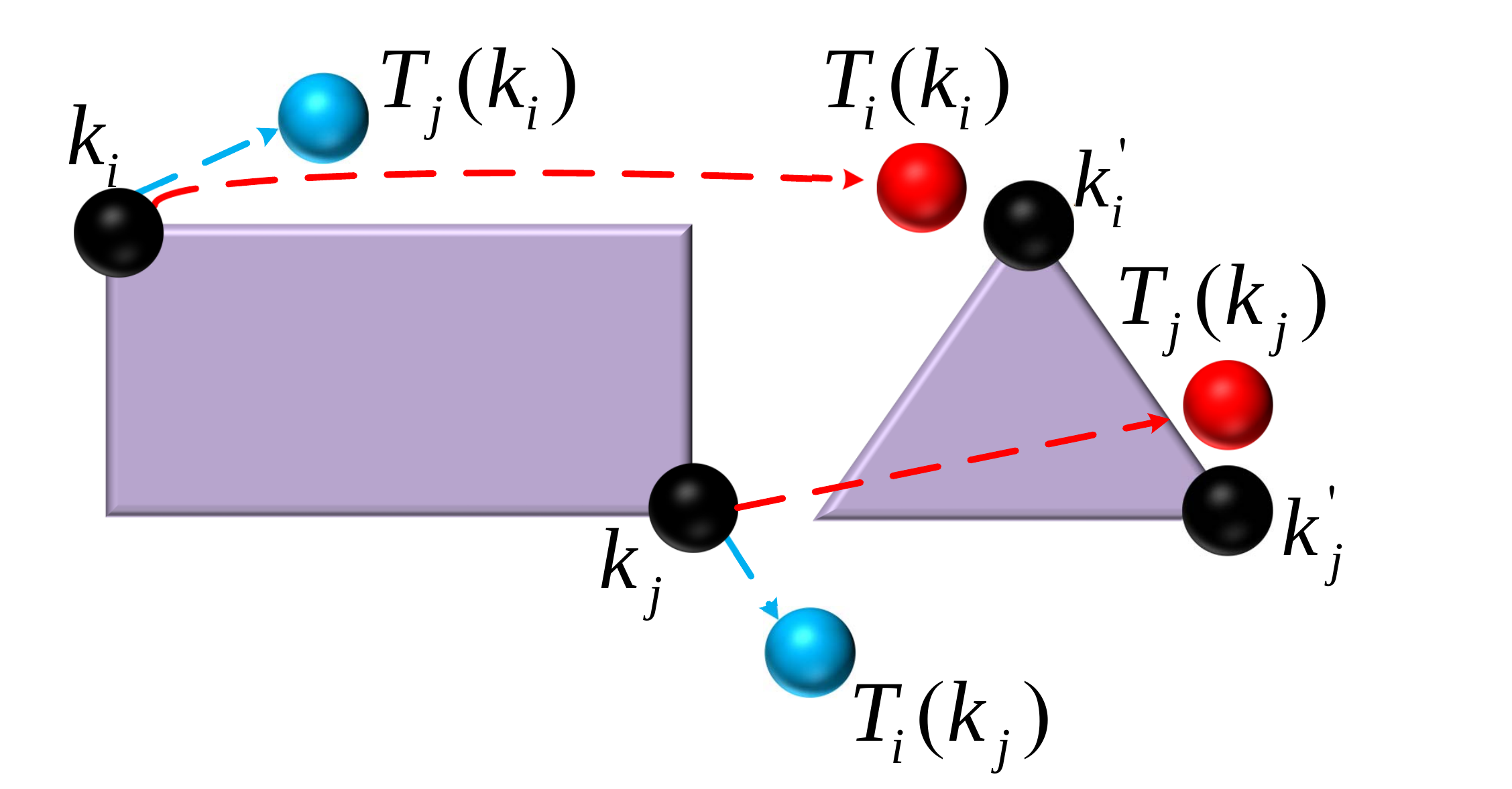}}
	\caption{Illustrations of geometric compatibility of two inliers (a) and two outliers (b), where the yellow dots indicate the keypoints of inliers, the black dots represent the keypoints of outliers, and the red and blue dots show the keypoint positions respectively projected by two local transformations. The geometric structures around inliers are similar (two quadrangles from different viewpoints), resulting in a closer Euclidean distance between projected keypoints ($({{T}_{i}}({\mathbf{k}_{i}}), {{T}_{j}}({\mathbf{k}_{i}}))$ as an example), yet the geometric structures around outliers are varied (quadrangle and triangle), leading to a farther Euclidean distance.}
	\label{fig:payoff}
\end{figure}
\textbf{Geometric compatibility.  } Inspired by the recent work~\cite{ma2019locality}, geometric structures in the neighborhood of inliers tent to be homogeneous as shown in Fig.~\ref{fig:payoff} (a) (two quadrangles from different viewpoints), which results in consistent local transformations. By contrast, the variation of geometric structures around outliers can be large and irregular as shown in Fig.~\ref{fig:payoff} (b) (quadrangle and triangle), leading to inconsistent local transformations. Based on the above observations, we suggest the use of Euclidean distance between the keypoint positions projected by the pair of local transformations around two correspondences as a measure of geometric compatibility. In other words, from the perspective of local geometric variations, we can define geometric compatibility as  
\begin{equation}\label{eq:LRF4}
{P_{ij}^{geo}}={{e}^{-\frac{\left\| {{T}_{i}}({\mathbf{k}_{i}})-{{T}_{j}}({\mathbf{k}_{i}}) \right\| + \left\| {{T}_{i}}({\mathbf{k}_{j}})-{{T}_{j}}({\mathbf{k}_{j}}) \right\|}{\sigma}}},
\end{equation}
where $||\cdot||$ represents the $L_2$ norm, ${T}_{j}(\mathbf{k}_i)$  is the projected position of $\mathbf{k}_i$ (an exemplar keypoint) through local transformation ${T}_{j}$ and calculated by (${T}_{i}(\mathbf{k}_i)$ in the same way)
\begin{equation}\label{eq:LRF30}
{T}_j({\mathbf{k}_i})=\rho \left( \left[ \begin{matrix}
\mathbf{A}_j & \mathbf{k}_j  \\
\mathbf{0} & 1  \\
\end{matrix} \right]\cdot\left[ \begin{matrix}
\mathbf{k}_i  \\
1  \\
\end{matrix} \right] \right),
\end{equation}
where $\rho([a_1\;a_2\;a_3]^{T})=[a_1/a_3\;a_2/a_3]^{T}$, $\mathbf{A}_j$ being the $2\times2$ affine information arrond $\mathbf{k}_j$, and $\sigma$ is a scale coefficient. To obtain $\mathbf{A}_j$, one might use the off-the-shelf keypoing detection method (e.g., Hessian-affine detector~\cite{Mikolajczyk2004Hessian}). 
\\
\\
\textbf{Descriptive compatibility.  } Consider a salient keypoint in the real world; the projections of this keypoint onto two imaging planes ($I,I^{'}$) - two local descriptive features - should be similar. A straightforward approach of measuring the similarity of descriptive features  (e.g., SIFT descriptor~\cite{lowe2004distinctive}) is to calculate their Euclidean distance between $(\mathbf{f}_{i}, \mathbf{f}_{i}^{'})$ of $c_i$. However, it is well known that $L_2$-norm is not robust to outliers and easily confused by nuisances such as nonuniform illumination, motion blur, and viewpoint variations. A more robust strategy is to use relative (instead of absolute) difference between descriptive features. For example, the so-called divisive normalization \cite{lyu2008nonlinear} strategy shows improved robustness over conventional $L_2$-norm.

Here we have adopted the ratio test as an alternative metric whose robustness has been shown in ~\cite{lowe2004distinctive}. The ratio-based descriptive compatibility is defined by
\begin{equation}\label{eq:LR15}
{{r}_{i}}=\frac{\left\| {\mathbf{f}_{i}}-\mathbf{f}_{nearest}^{'} \right\|}{\left\| {\mathbf{f}_{i}}-\mathbf{f}_{nearest2}^{'} \right\|},
\end{equation}
where $\mathbf{f}_{nearest}^{'}$ is the descriptor vector in $I^{'}$ closest to $\mathbf{f}_{i}$ and $\mathbf{f}_{nearest2}^{'}$ is the second closest descriptor vector. A credible correspondence is expected to achieve a significant distinctiveness between the closest match and the second closest match, resulting in a smaller ratio $r_i$. To measure the compatibility of two correspondences from the perspective of local feature embedding, we expect both two correspondences perform prominent distinctiveness if they are consistent. Therefore, we can define the descriptive compatibility payoff term by
\begin{equation}\label{eq:LRF6}
P_{ij}^{des}={{e}^{-\frac{\max ({{r}_{i}},{{r}_{j}})}{\alpha }}},
\end{equation}
where $\alpha$ is a scale coefficient.

\textbf{Evolutionary Stable Strategy.  } With the definition of the-above designed payoff functions, the popularity of all players can be iteratively updated by the evolutionary stable strategy (ESS)~\cite{weibull1997evolutionary} algorithm as 
\begin{equation}\label{eq:LRF10}
{{{q}_{i}}(k+1)={{q}_{i}}(k)\frac{{{(\mathbf{M}\mathbf{q}(k)})_{i}}}{\mathbf{q}{{(k)^{T}}}\mathbf{M}\mathbf{q}(k)}},
\end{equation}
where $\mathbf{q}$ is the popularity vector of all players and $\mathbf{M}$ is the payoff matrix generated by 
\begin{equation}\label{eq:LRF11}
{{m}_{ij}}=\left\{ \begin{aligned}
& {{P}_{ij}},\text{ if } i\ne j \\ 
& 0,\text{ otherwise} \\ 
\end{aligned} \right.,
\end{equation}
where $m_{ij}\in{\mathbf{M}}$. As the game going on, the popularity of players who acquire larger payoffs from the other players is significantly higher, which indicates their selections are comparable to the majority of other correspondences. Quantitatively, $c_i$ is determined as a correct match if the corresponding $q_i$ is higher than an adaptive threshold calculated by the OTSU~\cite{otsu1979threshold} algorithm.

\subsection{Iterative Clustering}
\begin{figure}[t]
	\centering 
	\includegraphics[width=1\linewidth]{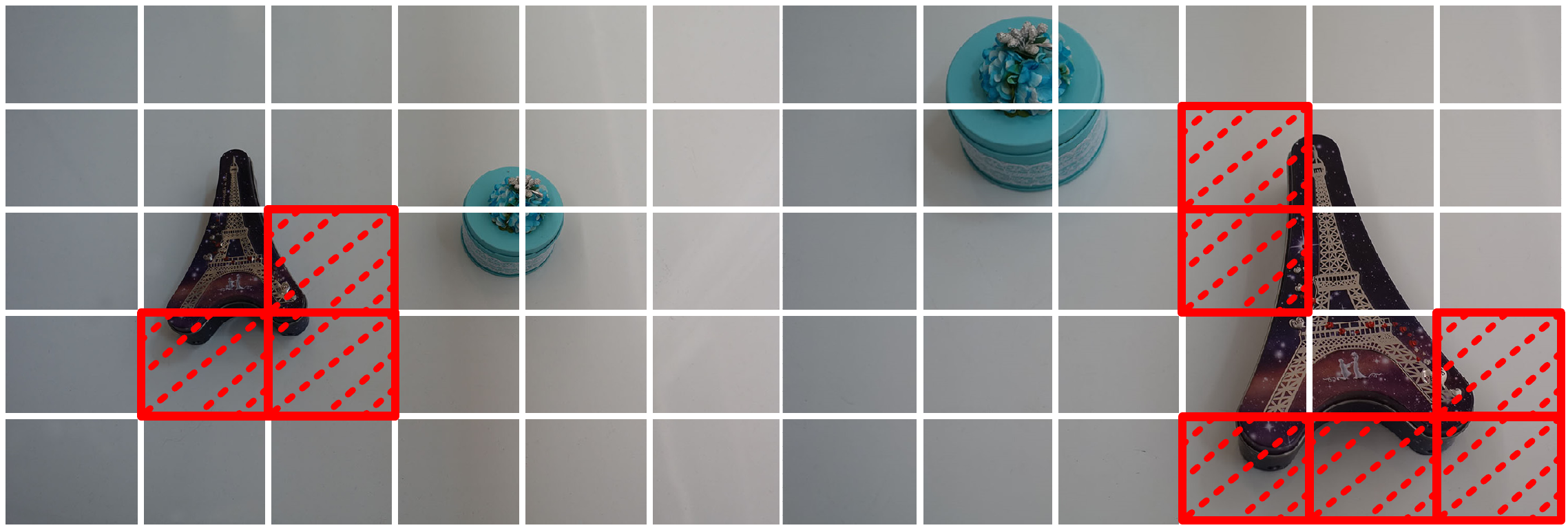}
	\caption{Motivation for iterative clustering. The red boxes show corners or edges of the object, where few correspondences can be found and tend to be eliminated in the block matching step.}
	\label{fig:moti}
\end{figure}
Local non-cooperative games produce some candidate correspondences for multi-consistency matching. However, they are not optimized and need to be further refined/clustered for the following reasons. First, grid-based block matching is an efficient yet approximate process. As illustrated in Fig.~\ref{fig:moti}, the red blocks contain corners or edges of the object, and fewer correspondences are located in these areas, which are prone to be eliminated by block matching after thresholding. Second, GTM is known to suffer from high miss detection or false rejection rate (i.e., poor recall performance) because many correct matches are falsely rejected by local games ~\cite{Rodola2014Elastic}. As shown in Fig.~\ref{fig:way}, the correspondences after playing local games are overwhelmingly consistent but sparse. Third,  local games cannot resolve the ambiguity underlying multiple consistencies. In other words, the collection of correspondences still need to be classified and assigned to different consistency classes.

To address these issues, we propose an iterative clustering process for simultaneously recovering falsely-rejected inliers and classifying different consistencies. Intuitively, our approach can be interpreted as a generalized clustering in the space of matched correspondences. Specifically, our iterative clustering method consists of four steps. First, we recompute the payoff matrix $\mathbf{M}^{*}$ for all selected candidates as
\begin{equation}\label{eq:LRF18}
{{m}_{ij}^{*}}=\left\{ \begin{aligned}
& {{P}_{ij}},\text{ if } i\ne j \\ 
& 0,\text{ otherwise} \\ 
\end{aligned} \right., c_i\in{\mathcal{C}^{*}},
\end{equation}
where $m_{ij}^{*}\in{\mathbf{M}^{*}}$ and $\mathcal{C}^{*}$ is the set of candidates. Second, we find out the most currently consistent pair of correspondences as an anchor, which corresponds to the maximum element in ${\mathbf{M}^{*}}$. Third, we search for  and cluster the other correspondences consistent with the anchor by comparing the corresponding elements with a threshold defined as 
\begin{equation}\label{eq:LRF19}
\tau =\frac{\max ({{m}_{ij}^{*}})+\min ({{m}_{ij}^{*}})}{2}.
\end{equation}
Fourth, the clustered correspondences are removed by set elements in the corresponding row and column to be zero. The steps 2-4 can be iterated until the size of clustered subset is lower than a predefined threshold (4 in our experiment).

To quantitatively evaluate the consistencies represented by parametric transformations, we perform RANSAC (other parametric methods can be used as well) within each cluster and simultaneously calculate a set of parametric transformations (homography matrices in our approach) as $\mathcal{H}=\{{\mathbf{H}_i}, i=1, 2, 3..., L\}$ ($L$ is the number of clusters). Finally, $\mathcal{H}$ is employed to check the initial correspondences and recover the falsely eliminated inliers by computing re-projected errors as

\begin{equation}\label{eq:LRF9}
E_{j}^{i}=\left\| \rho \left( {\mathbf{H}_{i}}\left[ \begin{matrix}
\mathbf{k}_{j}  \\
1  \\
\end{matrix} \right] \right)-{\mathbf{k}_{j}^{'}} \right\|,
\end{equation}\label{subsec:enrich}
where $E_j^{i}\in{\mathcal{E}_j}$, $\mathbf{H}_i\in{\mathcal{H}}$ and $(\mathbf{k}_j,\mathbf{k}_j^{'})  \in{c_j}$.
$c_j\in\mathcal{C}$ is determined as an inlier if the minimum element in ${\mathcal{E}_j}$ is lower than a threshold $t$ ($5$ in default).

\section{Performance Evaluation}
\label{sec:metric}
For single consistency feature matching, precision (P), recall (R), and F-measure (F) are commonly used for performance measurement~\cite{lin2014bilateral, bian2017gms, ma2019locality}. However, these metrics are not appropriate in case of multi-consistency feature matching. As demonstrated in Fig.~\ref{fig:metric}, due to the difference of geometric structures or image textures, the spatial distribution of correspondences tends to be vary across different regions. Therefore, there is a large gap between underlying consistencies associated with different moving objects especially from the perspective of saliency. For example, less salient consistencies that contain fewer correspondences are often prone to be eliminated (e.g., the consistencies highlighted by red color in Fig.~\ref{fig:metric}), but its impact on the actual performance measured by P, R, and F will be insignificant because the eliminated inliers only make up a small portion of correspondences. 

To make up the above deficiency, we propose to use three new metrics, -i.e., weighted-precision (W-P), weighted-recall (W-R), and weighted-F-measure (W-F) that are more appropriate for evaluating the performance of multi-consistency matching. The key new insight is to introduce the idea of weighting while distinguishing correspondences among different consistencies. More specifically, each match is weighted according to the number of correspondences within the consistency it belongs to. That is, the weight is negatively correlated with the number of incliers consistent with the associated model (homography H) - i.e.,

\begin{equation}
{w_{inlier}^{i}}=\frac{{{e}^{-{{{N}_{i}}}/{{{N}_{inlier}}}\;}}}{\sum\limits_{i=1}^{K}{{{e}^{^{-{{{N}_{i}}}/{{{N}_{inlier}}}\;}}}}},
\end{equation}

where $w_i$ is the weight of an inlier belonging to the $i$-th consistency $\mathbf{H}_i$, $N_i$ is the number of inliers consistent with $\mathbf{H}_i$, and $N_{inlier}$ is the total number of inliers. For outliers, the weight is calculated by

\begin{equation}
{{w}_{outlier}}=\max (\{w_{inlier}^{i},i=1,2,3,...,K\}).
\end{equation}
Note that we choose the maximum operation for the purpose of magnifying the penalty of outliers on performance metrics.
Therefore, W-P, W-R, and W-F are respectively defined by

\begin{equation}
\text{W-P}=\frac{{{W}_{TP}}}{{{W}_{TP}}+{{W}_{FP}}},
\end{equation}

\begin{equation}
\text{W-R}=\frac{{{W}_{TP}}}{{{W}_{TP}}+{{W}_{FN}}},
\end{equation}

\begin{equation}
\text{W-F}=\frac{\text{W-P}\times \text{W-R}}{\text{W-P}+\text{W-R}},
\end{equation}
where $W$ is the sum of all $w$'s (including $w_{inlier}$ and $w_{outlier}$), $W_{TP}$, $W_{FP}$, and $W_{FN}$ respectively represent True Positive, False Positive, and False Negative results (i.e., $\mathcal{C}_{TP}$, $\mathcal{C}_{FP}$, and $\mathcal{C}_{FN}$) generated by the evaluation method.

\section{Experiments}
\label{sec:exper}
In this section, we will elaborate on our benchmark~\ref{sec:ben} including the datasets (\ref{sec:dataset}) and the experimental setup (\ref{sec:setup}), present the quantitative (\ref{sec:quan}) and qualitative (\ref{sec:qual}) experimental results, and conduct some analysis (\ref{sec:ana}) including the analysis about payoff function (\ref{sec:bin}) and ablation study (\ref{sec:abl}) to better illustrate the mechanism behind ic-GTM.

\subsection{Benchmark}
\label{sec:ben}

\subsubsection{Datasets}
\label{sec:dataset}
\begin{figure}[t]
	\centering
	\subfigure[Scene-1]
	{ \includegraphics[width=1.0\linewidth]{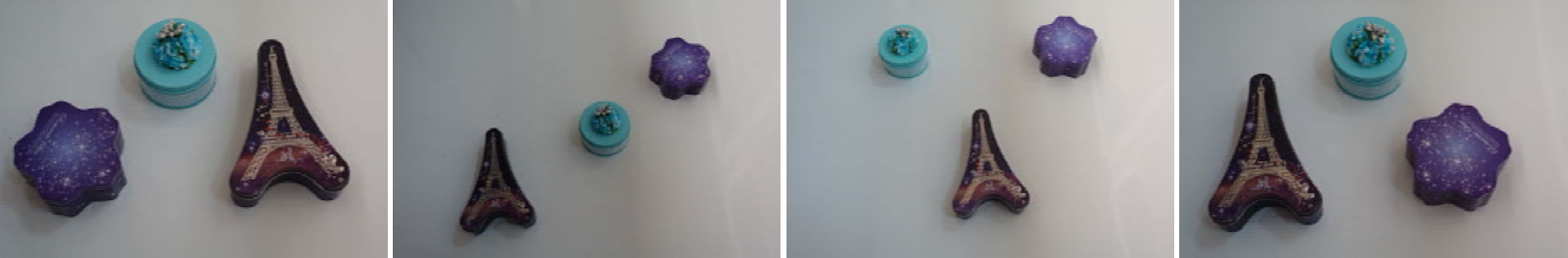}}
	\subfigure[Scene-2]
	{ \includegraphics[width=1.0\linewidth]{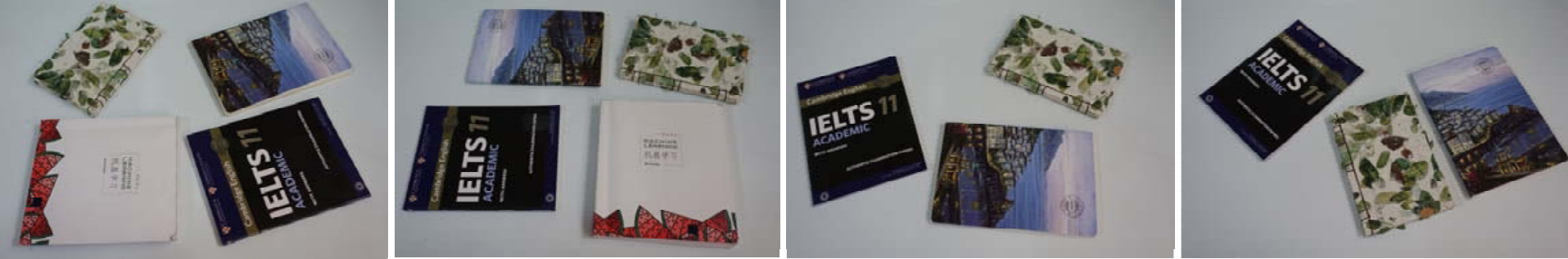}}
	\subfigure[Scene-3]
	{ \includegraphics[width=1.0\linewidth]{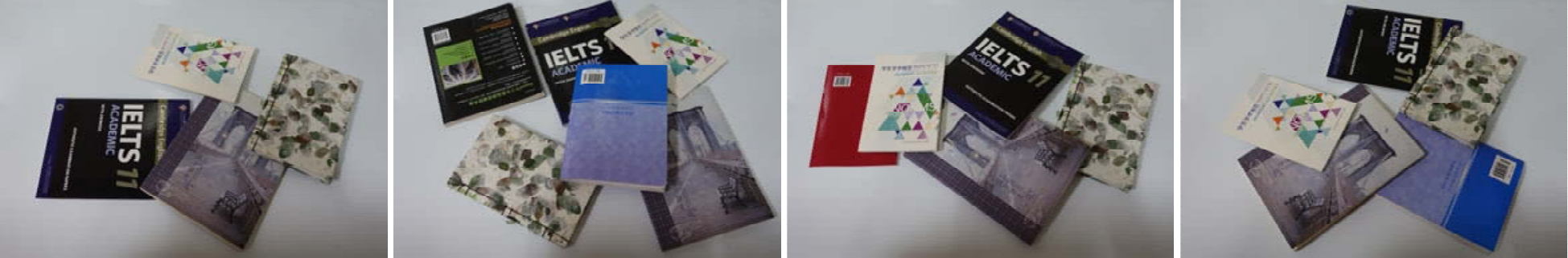}}
	\subfigure[AdelaideRMF]
	{ \includegraphics[width=1.0\linewidth]{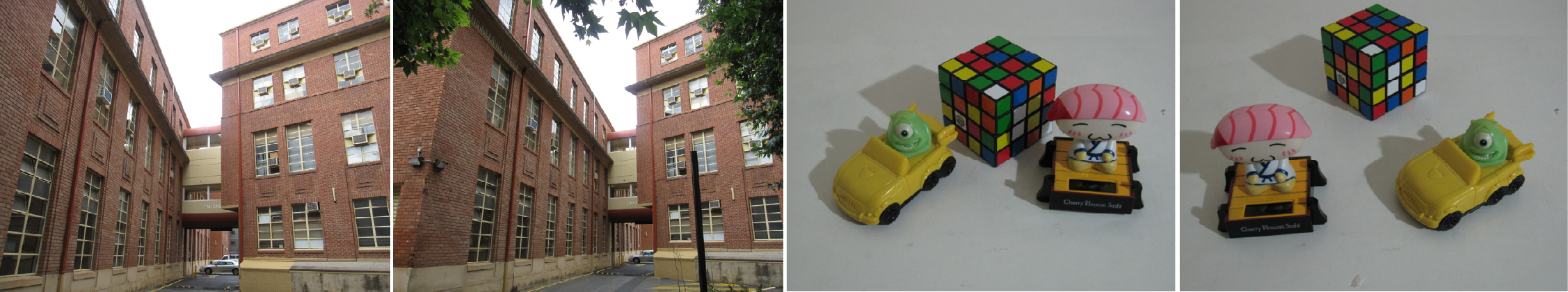}}
	\subfigure[VGG]
	{ \includegraphics[width=1.0\linewidth]{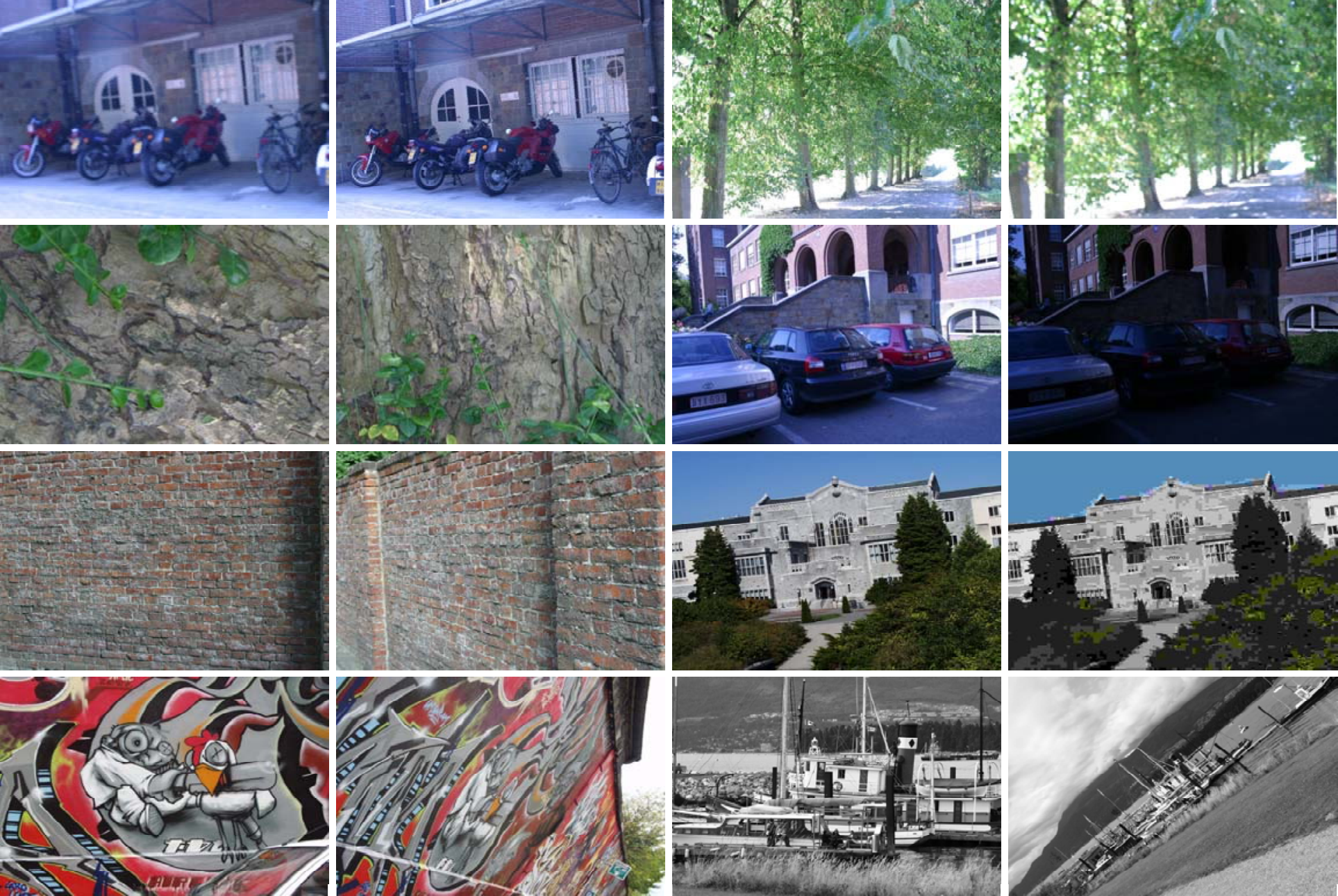}}
	\caption{Examples of the datasets. (a) (b) (c) are three dynamic scenes in our collected dataset, and (d) (e) are two public datasets contain multiple consistencies and single consistency respectively.}
	\label{fig:fig2}
\end{figure}

\begin{figure}[!t]
	\centering 
	\includegraphics[width=1\linewidth]{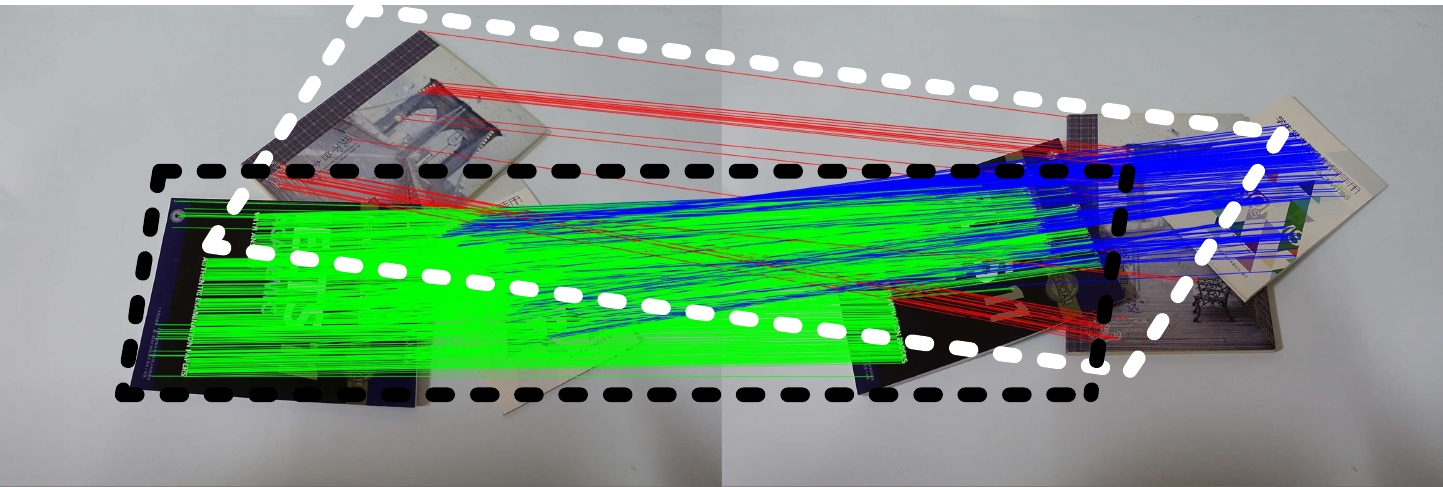}
	\caption{Difficulty with traditional P, R, F in the case of multi-consistency feature matching. The number of inliers in the white box is significantly lower than the one in the black box. However, the missing of red consistency will not have a significant impact on the performance measured by P, R, and F.}
	\label{fig:metric}
\end{figure}

In the field of correspondence selection, most existing datasets have only considered the cases of single consistency and have not covered dynamic scenes which contain multiple consistencies due to the presence of moving objects/camera. To fill the gap of multi-consistency evaluation, we have set up a dataset consisting of three dynamic scenes with varying challenges, -i.e., translation, rotation, clutter, and occlusion. The ground truth in our dataset is a subset of manually labelled correspondences (inliers). To make our benchmark more comprehensive, we have also included a classical public dataset, -i.e., AdelaideRMF~\cite{wong2011dynamic}, in which each image pair includes multiple consistencies among different structures. Meantime, we still employ VGG dataset~\cite{mikolajczyk2005comparison} to evaluate the generalization of ic-GTM for single consistency feature matching. Some examples and characteristics of those datasets in our benchmark are shown in Fig.~\ref{fig:fig2} and Table~\ref{tab:dataset}.
\begin{table*}[t]
	\renewcommand{\arraystretch}{1}
	\caption{Characteristics of the dataset.}
	\label{tab:dataset}
	\centering
	\begin{tabular}	{lcccc}
		\hline
		Dataset & Challenges & Ground truth & \# Image pairs 	\\
		\hline
		Scene-1 & Translation and zoom & Manually labeled inliers & 15	\\
		\hline
		Scene-2 &  Translation and rotation & Manually labeled inliers & 15 \\
		\hline
		Scene-3 & Translation, rotation, clutter, and occlusion & Manually labeled inliers & 15 \\
		\hline
		AdelaideRMF~\cite{wong2011dynamic} & Multiple structures and viewpoint change & Manually labeled inliers & 38 \\
		\hline
		VGG~\cite{mikolajczyk2005comparison} & Zoom, rotation, blur, light change, viewpoint change, and JPEG compression & Homography matrix & 40 \\
		\hline
	\end{tabular} 
\end{table*}

\subsubsection{Experimental setup}
\label{sec:setup}
We have evaluated ic-GTM along with seven other competing methods including RANSAC~\cite{fischler1981random}, GTM~\cite{albarelli2012imposing}, Multi-GS~\cite{chin2010accelerated}, USAC~\cite{Raguram2013USAC}, VFC~\cite{Ma2014Robust}, GMS~\cite{bian2017gms} and LPM~\cite{ma2019locality}. The evaluation metrics are weighted-precision (W-P), weighted-recall (W-R), weighted-F-measure (W-F), and efficiency (T) for our dataset. F-measure  (F) is also used in our dataset to verify the superiority of our metrics. For single consistency, we have used precision (P), recall (R), F-measure (F), and efficiency (T) in VGG. Each image is divided into $5\times 5$ blocks, and the initial correspondence set is generated by brute-force matching~\cite{lowe2004distinctive} with the combination of Hessian-affine detector~\cite{Mikolajczyk2004Hessian} and SIFT descriptor~\cite{lowe2004distinctive}. Notably, since only SIFT detector is provided in AdelaideRMF and the affine information in Eq.~\ref{eq:LRF30} is unavailable, only quantitative results are shown on this dataset.

\subsection{Quantitative Results}
\label{sec:quan}

\subsubsection{Single consistency}
\begin{table*}[!t]
	\renewcommand{\arraystretch}{1.3}
	\caption{Quantitative results on VGG dataset measured by precision, recall, F-measure, and time.}
	\label{tab:single}
	\centering
	\begin{tabular}{cp{1.8cm}<{\centering}p{1cm}<{\centering}p{1cm}<{\centering}p{1.2cm}<{\centering}p{1cm}<{\centering}p{1cm}<{\centering}p{1cm}<{\centering}p{1cm}<{\centering}p{1.4cm}<{\centering}}
		\hline
		& & RANSAC & GTM & Multi-GS & USAC & VFC & GMS & LPM & ic-GTM \\
		\hline
		\hline
		Case-1 & Precision (\%) & 79.79 & 43.22 & 13.64 & \textbf{80.57} & 67.19 & 63.61 & 42.5 & 78.50\\
		(zoom rotation) & Recall (\%) & 91.68 & 79.6 & 7.39 & 97.49 & 86.11 & 11.45 & 83.54 & \textbf{99.29}  \\
		& F-measure (\%) & 84.86 & 53.69 & 9.2 & \textbf{86.32} & 74.27 & 18.46 & 54.75 & 85.38  \\
		& Time (s) & 0.03 & 0.43 & 2.30 & 0.02 & 0.005 & \textbf{0.001} & 0.01 & 0.03\\
		\hline
		Case-2 & Precision (\%) & 37.91 & \textbf{67.00} & 15.46 & 38.90 & 29.44 & 41.47 & 27.73 & 57.63 	\\
		(blur) & Recall (\%) & 39.16 & 56.74 & 8.62 & 40.00 & 51.41 & 50.45 & 54.75 & \textbf{73.35}   \\
		& F-measure (\%) & 38.32 & 61.12 & 10.00 & 39.44 & 35.27 & 45.54 & 35.86 & \textbf{64.09} 	\\
		& Time (s) & 0.06 & 12.05 & 4.70 & 0.75 & 0.02 & \textbf{0.001} & 0.02 & 0.25 \\
		\hline
		Case-3 & Precision (\%) & 72.40 & 44.92 & 19.60 & 57.43 & 49.38 & 58.57 & 44.59 & \textbf{71.92} 	\\
		(zoom rotation) & Recall (\%) & 77.50 & 52.16 & 14.79 & 57.06 & 99.22 & 57.21 & 76.43 & \textbf{99.90}   \\
		& F-measure (\%) & 73.52 & 44.83 & 14.00 & 57.23 & 61.91 & 56.35 & 55.10 & \textbf{82.40}  	\\
		& Time (s) & 0.05 & 54.29 & 7.70 & 0.52 & 0.04 & \textbf{0.002} & 0.05 & 0.92\\
		\hline	
		Case-4 & Precision (\%) & 64.08 & 50.94 & 25.86 & 52.75 & 57.86 & 57.05 & 45.08 & \textbf{69.30} 	\\
		(viewpoint change) & Recall (\%) & 57.74 & 68.55 & 27.76 & 53.50 & \textbf{97.08} & 75.52 & 83.87 & 84.94   \\
		& F-measure (\%) & 56.67 & 55.69 & 25.00 & 53.12 & 71.23 & 64.55 & 56.56 & \textbf{74.68}  	\\
		& Time (s) & 0.04 & 43.50 & 6.90 & 0.53 & 0.04 & \textbf{0.002} & 0.04 & 0.84\\
		\hline	
		Case-5 & Precision (\%) & 88.75 & 68.90 & 31.65 & \textbf{96.26} & 71.99 & 64.89 & 57.65 & 84.32 	\\
		(light change) & Recall (\%) & 80.94 & 80.35 & 26.88 & 99.91 & 100 & 87.95 & 84.46 & \textbf{100}   \\
		& F-measure (\%) & 81.25 & 73.94 & 26.00 & \textbf{98.05} & 82.49 & 74.37 & 67.90 & 91.38  	\\
		& Time (s) & 0.04 & 53.08 & 8.10 & 0.10 & 0.05 & 0.001 & 0.05 & 2.29\\
		\hline	
		Case-6 & Precision (\%) & 49.85 & 33.45 & 5.62 & 35.88 & 31.18 & 57.10 & 26.72 & \textbf{58.79} 	\\
		(blur) & Recall (\%) & 21.19 & 39.10 & 2.21 & 39.65 & 40.00 & 47.00 & 66.81 & \textbf{87.31}   \\
		& F-measure (\%) & 23.74 & 34.29 & 2.40 & 37.65 & 35.02 & 50.80 & 35.82 & \textbf{68.95}  	\\
		& Time (s) & 0.08 & 50.70 & 7.40 & 0.74 & 0.04 & 0.002 & 0.04 & 0.42\\
		\hline	
		Case-7 & Precision (\%) & 93.04 & 80.66 & 49.15 & \textbf{97.57} & 89.48 & 79.87 & 75.87 & 91.45 	\\
		(JPEG compression) & Recall (\%) & 98.48 & 94.42 & 58.01 & 100 & 100 & 96.70 & 93.38 & \textbf{100}   \\
		& F-measure (\%) & 95.63 & 86.88 & 53.00 & \textbf{98.77} & 94.26 & 87.25 & 83.43 & 95.46  	\\
		& Time (s) & 0.009 & 50.70 & 7.90 & 0.06 & 0.04 & 0.002 & 0.05 & 2.86\\
		\hline	
		Case-8 & Precision (\%) & 74.59 & 72.05 & 25.43 & 76.71 & 72.40 & 80.86 & 62.67 & \textbf{84.60} 	\\
		(viewpoint change) & Recall (\%) & 72.49 & 73.42 & 8.60 & 77.66 & 75.74 & 76.08 & 69.64 & \textbf{99.28}   \\
		& F-measure (\%) & 48.68 & 39.98 & 26.24 & 63.23 & 44.86 & 74.04 & 77.80 & \textbf{90.77}  	\\
		& Time (s) & 0.04 & 50.74 & 7.30 & 0.27 & 0.05 & \textbf{0.002} & 0.05 & 0.60\\
		\hline	
		average & Precision (\%) & 70.05 & 23.31 & 57.64 & 67.01 & 58.61 & 62.96 & 47.85 & \textbf{74.57} 	\\
		& Recall (\%) & 67.40 & 18.87 & 68.88 & 70.78 & 81.67 & 62.42 & 78.14 & \textbf{93.01}   \\
		& F-measure (\%) & 65.91 & 18.54 & 60.48 & 68.53 & 66.27 & 59.17 & 57.38 & \textbf{81.64}	\\
		& Time (s) & 0.05 & 6.54 & 39.44 & 0.37 & 0.04 & \textbf{0.002} & 0.04 & 1.02\\
		\hline
	\end{tabular}
\end{table*}
Although the focus of this work is multi-consistency matching, the proposed ic-GTM can be easily generalized for single-consistency matching. From a performance evaluation perspective, we think it is worth including the comparison for static scenes or single consistency feature matching as the starting point. As shown in Table~\ref{tab:single}, ic-GTM achieves superior average performance, -i.e., $74.57\%$, $93.01\%$, and $81.64\%$, which outperform all other competing algorithms by a large margin. Moreover, ic-GTM has achieved promising results in all eight cases with varying challenges from geometric structure diversity to image quality variations, which confirms the robustness property of ic-GTM. Meantime, it should be noted that the performance of ic-GTM in case-2 and case-6 in the presence of blurred images is significantly lower than other cases. This is because the descriptive compatibility item in Eq.~\ref{eq:LRF3} is confused by severe degradation of image qualities when blur occurs. %The robustness of ic-GTM towards the image quality variations needs to be further improved.

\subsubsection{Multiple consistencies}

\begin{figure*}[!t]
	\centering 
	\includegraphics[width=1\linewidth]{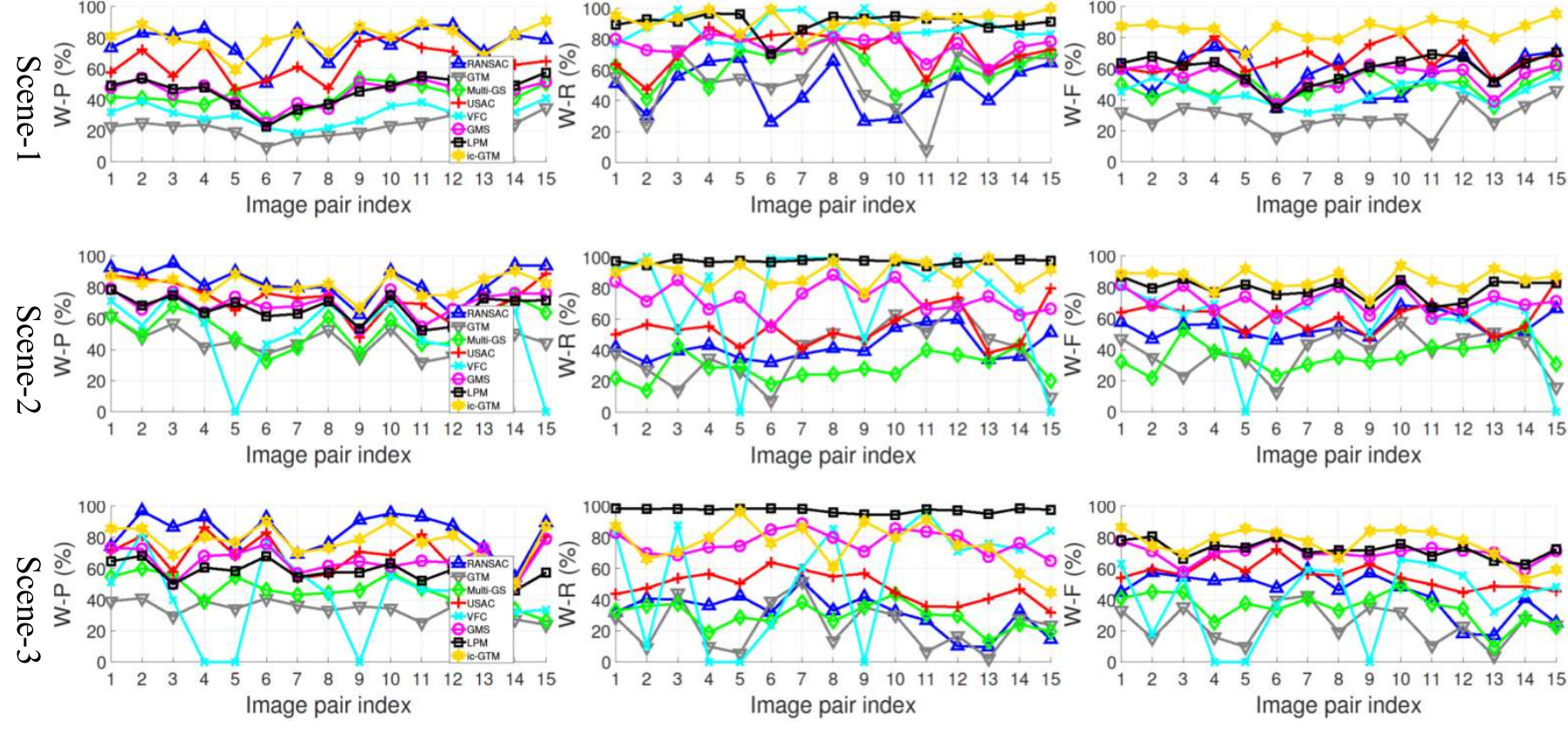}
	\caption{Performance on each image pair in scene-1, scene-2, and scene-3 of our dataset evaluated by weighted-precision, weighted-recall, and weighted-F-measure.}
	\label{fig:multi}
\end{figure*}

\begin{table*}[!t]
\renewcommand{\arraystretch}{1.3}
\caption{Quantitative results on our dataset measured by weighted-precision, weighted-recall, weighted-F-measure, F-measure, and time.}
\label{tab:multi}
\centering
\begin{tabular}{lp{2.5cm}<{\centering}p{1.2cm}<{\centering}p{1.2cm}<{\centering}p{1.2cm}<{\centering}p{1.2cm}<{\centering}p{1.2cm}<{\centering}p{1.2cm}<{\centering}p{1.2cm}<{\centering}p{1.2cm}<{\centering}}
\hline
& & RANSAC & GTM & Multi-GS & USAC & VFC & GMS & LPM & ic-GTM \\
\hline
\hline
Scene-1 & W-P (\%) & 77.36 & 41.1 & 21.99 & 62.93 & 29.97 & 43.69 & 44.89 & \textbf{79.67} \\
& W-R (\%) & 48.15 & 62.00 & 53.03 & 72.84 & 87.22 & 75.24 & 90.62 & \textbf{92.11}   \\
& W-F (\%) & 57.93 & 48.40 & 29.18 & 66.46 & 43.98 & 54.76 & 59.63 & \textbf{85.17}   \\
& F-measure (\%) & 65.77 & 62.40 & 38.07 & 78.17 & 56.99 & 64.19 & 69.59 & \textbf{90.50}  \\
& Time (s) & 0.03 & 5.23 & 3.59 & 0.13 & 0.01 & \textbf{0.001} & 0.02 & 0.53\\
\hline
Scene-2 & W-P (\%) & \textbf{83.13} & 53.41 & 46.05 & 72.56 & 50.83 & 69.49 & 66.64 & 81.25 	\\
& W-R (\%) & 42.26 & 28.61 & 38.26 & 54.40 & 74.00 & 73.27 & \textbf{97.20} & 89.60   \\
& W-F (\%) & 55.13 & 36.41 & 38.70 & 61.31 & 58.64 & 71.01 & 78.82 & \textbf{85.08} 	\\
& F-measure (\%) & 59.70 & 40.75 & 40.94 & 66.65 & 62.26 & 73.94 & 81.64 & \textbf{87.23}  \\
& Time (s) & 0.08 & 12.28 & 4.79 & 0.13 & 0.02 & \textbf{0.001} & 0.03 & 0.73\\
\hline
Scene-3 & W-P (\%) & \textbf{83.02} & 43.70 & 32.71 & 68.62 & 39.01 & 65.37 & 57.72 & 77.36 	\\
& W-R (\%) & 31.69 & 29.38 & 22.97 & 48.07 & 55.36 & 76.70 & \textbf{97.21} & 75.64  \\
& W-F (\%) & 44.20 & 34.76 & 24.49 & 55.56 & 39.81 & 70.02 & 72.22 & \textbf{75.62}  	\\
& F-measure (\%) & 48.68 & 39.98 & 26.73 & 62.85 & 44.86 & 74.04 & 77.80 & \textbf{79.23}  \\
& Time (s) & 0.07 & 13.55 & 5.27 & 1.06 & 0.02 & \textbf{0.001} & 0.07 & 0.70\\
\hline	
Average & W-P (\%) & \textbf{81.17} & 46.07 & 33.58 & 68.04 & 39.94 & 59.52 & 56.42 & 79.43	\\
& W-R (\%) & 40.7 & 40.00 & 38.09 & 58.44 & 72.19 & 75.07 & \textbf{95.01} & 85.78 \\
& W-F (\%) & 52.42 & 39.86 & 30.79 & 61.11 & 47.48 & 65.26 & 70.22 & \textbf{81.96}  \\
& F-measure (\%) & 58.05 & 47.71 & 35.04 & 69.13 & 54.70 & 70.72 & 76.34 & \textbf{85.65}	\\
& Time (s) & 0.07 & 13.55 & 5.27 & 1.06 & 0.02 & \textbf{0.001} & 0.07 & 0.65\\
\hline
\end{tabular}
\end{table*}

In Fig.~\ref{fig:multi}, we have plotted and compared the curves of three performance metrics (W-P, W-R and W-F) for three dynamic scenes. Different competing methods are represented by distinct color codes; it can be observed that the yellow color that represents the performance of ic-GTM is the best performing curve in most cases. Although the method of RANSAC (blue color) demonstrates strong W-P performance, its W-R and W-F performance dramatically fall behind. Such observation confirms that RANSAC is only good at discovering one kind of consistency, which is not appropriate for multi-consistency matching. LPM (black color) performs well in terms of W-R, but its W-P performance is disappointing. This is because LPM is not a selective method, with a relatively loose constraint. By contrast, ic-GTM is capable of striking an improved tradeoff between W-P and W-R, achieving the best overall W-F performance. The merits of local games and iterative clustering jointly contribute to its excellent performance. %Moreover, the curve of ic-GTM fluctuates significantly less than VFC, indicating that ic-GTM is more robust to the nuisances in different image pairs. 

The superiority of ic-GTM to other competing methods can also been verified from the quantitative results shown in Table~\ref{tab:multi}. We have compared the individual performance in each scene as well as the average performance on the entire dataset. As demonstrated in Table~\ref{tab:multi}, ic-GTM performs the best in terms of W-F and F, outperforming other approaches by a large margin. Besides, the computational efficiency of ic-GTM is improved at least an order of magnitude when compared with traditional GTM. This is because a large global payoff matrix in GTM is divided into some small matrices processed simultaneously by local games. Moreover, we note that the newly developed metrics (W-F as an example) seem more reasonable than traditional metrics (e.g., F). Taking RANSAC as an example, the W-F scores are remarkably lower than the corresponding F scores; this is because RANSAC tends to miss less salient consistencies. This deficiency is better reflected by the degradation of W-F performance than that of F-performance.

\subsection{Qualitative Results}
\label{sec:qual}

\begin{figure*}[!t]
	\centering
	{ \includegraphics[width=1.0\linewidth]{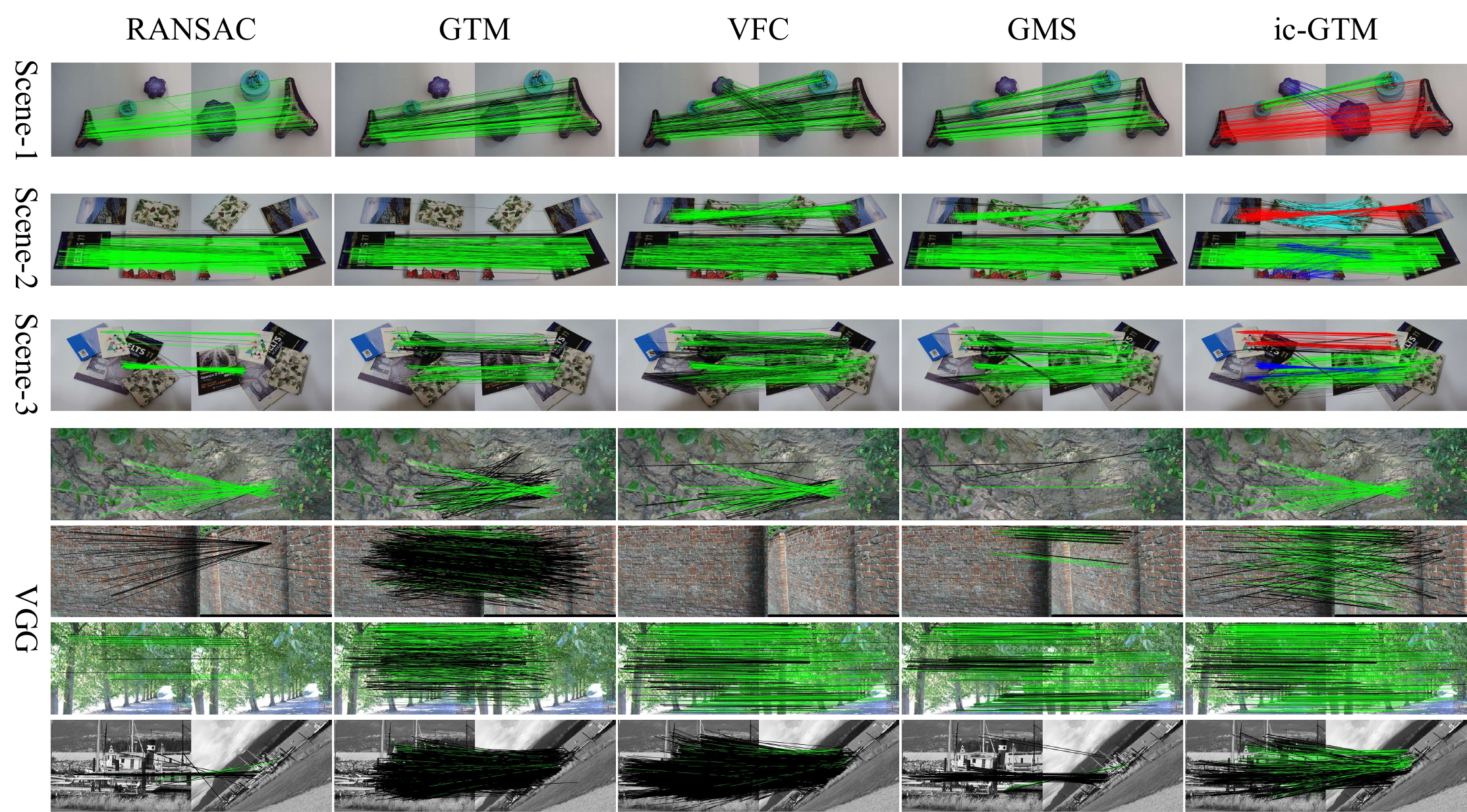}}
	\caption{Some exemplars of visual results in our datasets and VGG dataset. Notably, each color indicates a kind of consistency, yet black lines represent mismatches that are not correctly removed by the method. Only the green color is used for methods except ic-GTM because these methods cannot recognize multiple consistencies. }
	\label{fig:visual1}
\end{figure*}

\begin{figure*}[!t]
	\centering
	{ \includegraphics[width=1.0\linewidth]{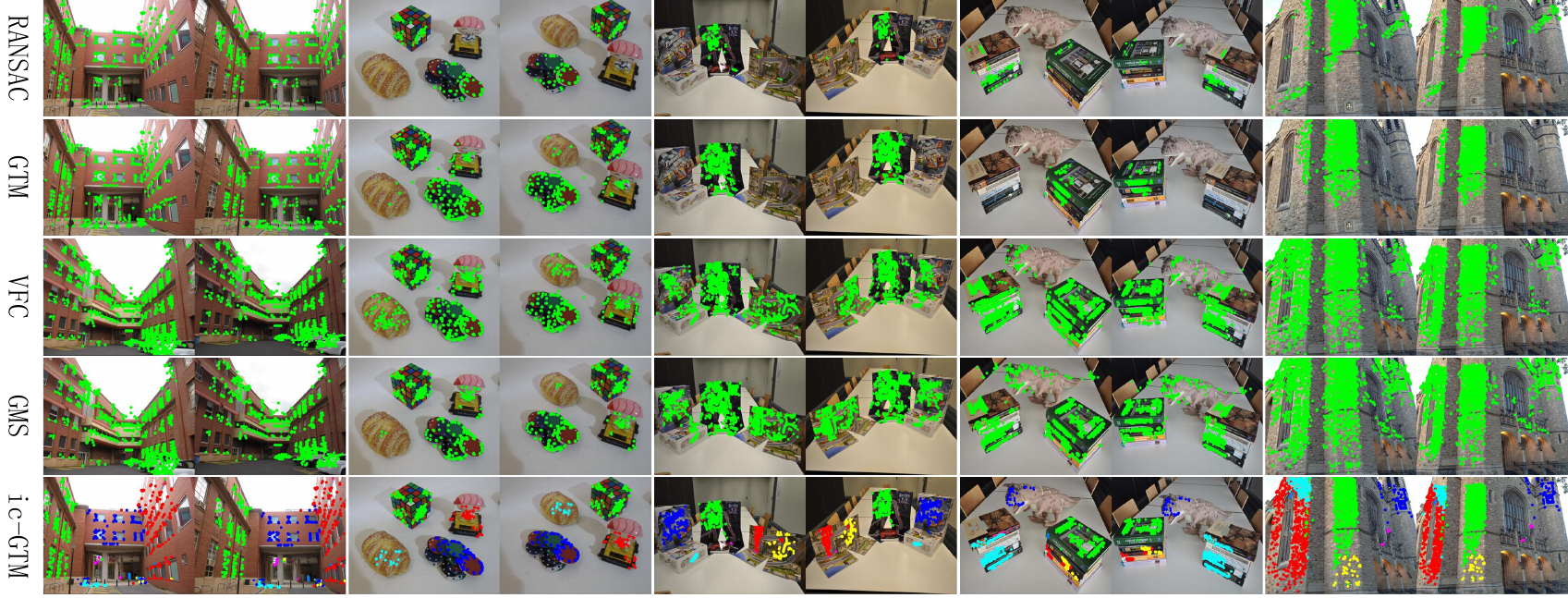}}
	\caption{Some exemplars of visual results in AdelaideRMF dataset. The dots are keypoint locations of the selected correspondences. The different colors are utilized to represent the different consistencies.}
	\label{fig:visual2}
\end{figure*}

We have also included visual comparison between ic-GTM and other competing methods on some exemplar scenes from our own dataset, VGG, and AdelaideRMF as shown in Fig.~\ref{fig:visual1} and Fig.~\ref{fig:visual2}. In Fig.~\ref{fig:visual1}, ic-GTM finds out most underlying consistencies on our dataset, which are highlighted by different colors (outliers or incorrect mismatches are represented by black color). Other methods such as RANSAC miss many correct matches and cannot recognize multiple consistencies in dynamic scenes. For VGG dataset, we have selected a few challenging cases, in which many other approaches are ineffective (e.g., dominated by black lines). By contrast, ic-GTM still achieves superior distinctiveness between the inlier and outlier in the presence of large scale viewpoint changes and blur. Note that the last row of Fig.~\ref{fig:visual1} contains large scale zoom and rotation (the most challenging case); ic-GTM works noticeably better than others but still suffer from many errors. 

In Fig.~\ref{fig:visual2}, we have used a different visualization methodology to compare different feature matching methods. Differently colored dots are used to indicate the keypoint positions of selected correspondences. Note that only ic-GTM produces multi-consistency matching results highlighted by different colors. In our experiment, SIFT detector provided in AdelaideRMF dataset is replaced by Hessian-affine detector in order to obtain the essential affine information required by ic-GTM. It is easy to see that only ic-GTM is capable of discovering the rich underlying consistencies characterized by either multiple planes in static scenes (e.g., building surfaces in the left and right two columns) or multiple moving objects in dynamic scenes (e.g., toys and books in the middle three columns). 

\subsection{Analysis}
\label{sec:ana}

\subsubsection{Payoff function}
There are several alternative choices of the payoff function in Eq.~\ref{eq:LRF3}. To compare their differences, we have evaluated the objective performance of ic-GTM using four different payoff functions on our dataset and VGG respectively. The comparison results are shown in Table~\ref{tab:payoff}. DES (descriptor) means the Euclidean distance between matched descriptor vectors which is defined as 
\begin{equation}
{{P}_{ij}}={{e}^{-\frac{\max (\left\| {\mathbf{f}_{i}}-\mathbf{f}_{i}^{'} \right\|,\left\| {\mathbf{f}_{j}}-\mathbf{f}_{j}^{'} \right\|)}{\beta }}}.
\end{equation}
DIS (distance) represents the first item of Eq.~\eqref{eq:LRF3}, R\_T (ratio test) corresponds to the second item of Eq.~\eqref{eq:LRF3}, and $R\_T + DIS$ denotes the sum of R\_T and DIS (both terms). We make the following observations from the reported comparison results. 

First, when compared with DES, R\_T achieves better performance in both multi-consistency and single-consistency scenarios, which confirms that the ratio test is more effective and robust than Euclidean distance. Meantime, DIS outperforms R\_T on our own dataset (dynamic scenes), but is worse than R\_T on the VGG dataset (static scenes). One possible explanation is that geometric compatibility is more effective than descriptive compatibility for less challenging static scenes in the absence of nonuniform illumination or motion  blur. 

Second, R\_T + DIS achieves the best performance on Scene-1, Scene-2, and Scene-3, outperforming both R\_T and DIS. Such result verifies that the combination of two payoff functions takes the advantages of both items, which demonstrates improved robustness for multi-consistency feature matching. However, we note that R\_T alone achieves the best results on VGG dataset, even surpassing R\_T + DIS. This shows the strategy of fusion has not been optimized in all scenarios; there is still room left to improve the choice of payoff function design (e.g., one might consider product-based instead of sum-based fusion).
\label{sec:bin}
\begin{table}[!t]
	\renewcommand{\arraystretch}{1}
	\caption{Comparison of four different payoff functions.}
	\label{tab:payoff}
	\centering
	\begin{tabular}	{p{1.0cm}<{\centering}p{1.2cm}<{\centering}p{0.6cm}<{\centering}p{0.6cm}<{\centering}p{0.6cm}<{\centering}p{1.5cm}<{\centering}p{1.3cm}<{\centering}}
		\hline
	    & & DES & R\_T & DIS & R\_T + DIS \\
		\hline
		\hline
		Scene-1 & W-P (\%) & 80.00 & \textbf{81.71} & 81.67 & 79.67 \\
		& W-R (\%) & 53.05 & 59.07 & 88.63 & \textbf{92.11}   \\
		& W-F (\%) & 59.44 & 66.17 & 84.65 & \textbf{85.17}  \\
		\hline
		Scene-2 & W-P (\%) & \textbf{83.69} & 83.56 & 82.39 & 81.25 	\\
		& W-R (\%) & 45.76 & 45.39 & 73.77 & \textbf{89.60}   \\
		& W-F (\%) & 57.69 & 57.17 & 76.78 & \textbf{85.08} 	\\
		\hline
		Scene-3 & W-P (\%) & \textbf{81.69} & 81.14 & 79.28 & 77.36 \\
		& W-R (\%) & 41.63 & 42.66 & 66.04 & \textbf{75.64}   \\
		& W-F (\%) & 53.56 & 54.15 & 70.28 & \textbf{75.62}   	\\
		\hline	
		\hline
		VGG~\cite{mikolajczyk2005comparison}  & P (\%) & 80.45 & \textbf{82.28} & 72.39 & 74.57 	\\
		& R (\%) & 91.82 & \textbf{94.50} & 87.05 & 93.01   \\
		& F (\%) & 85.00 & \textbf{87.09} & 77.97 & 81.64	\\
		\hline
	\end{tabular} 
\end{table}

\subsubsection{Ablation study}
\label{sec:abl}

\begin{table}[t]
	\renewcommand{\arraystretch}{1}
	\caption{Ablation study. }
	\label{tab:ablation}
	\centering
	\begin{tabular}	{cp{1.4cm}<{\centering}p{1.2cm}<{\centering}p{1.2cm}<{\centering}p{1.2cm}<{\centering}p{0cm}<{\centering}}
		\hline
		& & W-P (\%) &  W-R (\%) &  W-F (\%) \\
		\hline
		\hline
		Scene-1 & Without-EN & 76.20 & 28.56 & 40.73 \\
		& EN & \textbf{79.67} & \textbf{92.11} & \textbf{85.17}   \\
		\hline
		Scene-2 & Without-EN & 78.70 & 14.30 & 24.13 \\
		& EN & \textbf{81.25} & \textbf{89.60} & \textbf{85.08}   \\
		\hline
		Scene-3 & Without-EN & 65.02 & 15.16 & 24.26	\\
		& EN & \textbf{77.36} & \textbf{75.64} & \textbf{75.62}   \\
		\hline	
		& & P (\%) &  R (\%) &  F (\%) \\
		\hline
		\hline
		VGG~\cite{mikolajczyk2005comparison}  & Without-EN & 66.84 & 26.20 & 34.23 	\\
		& EN & \textbf{74.57} & \textbf{93.01} & \textbf{81.64}   \\
		\hline
	\end{tabular} 
\end{table}

Last but not the least, we report some ablation study result to further illustrate how the proposed ic-GTM method works. In particular, we want to shed some light to the relationship between playing local games (Sec. IIIC) and iterative clustering (Sec. IIID). As shown in Table~\ref{tab:ablation}, the implementation with iterative clustering surpasses the one without iterative clustering by a large margin in all cases. The performance gap is especially remarkable in terms of of W-R, which implies that a significant number of falsely-rejected inliers are recovered by the proposed iterative clustering process. Moreover, ic-GTM dramatically achieves better precision performance than conventional GTM because there is a double-check procedure reassuring the soundness of initial correspondences.

\section{Applications}
\label{sec:app}

\subsection{Dynamic Image Mosaicing}
\begin{figure}[!t]
	\centering 
	\subfigure[Misalignment artifacts of multi-structure images.]
	{ \includegraphics[width=0.9\linewidth]{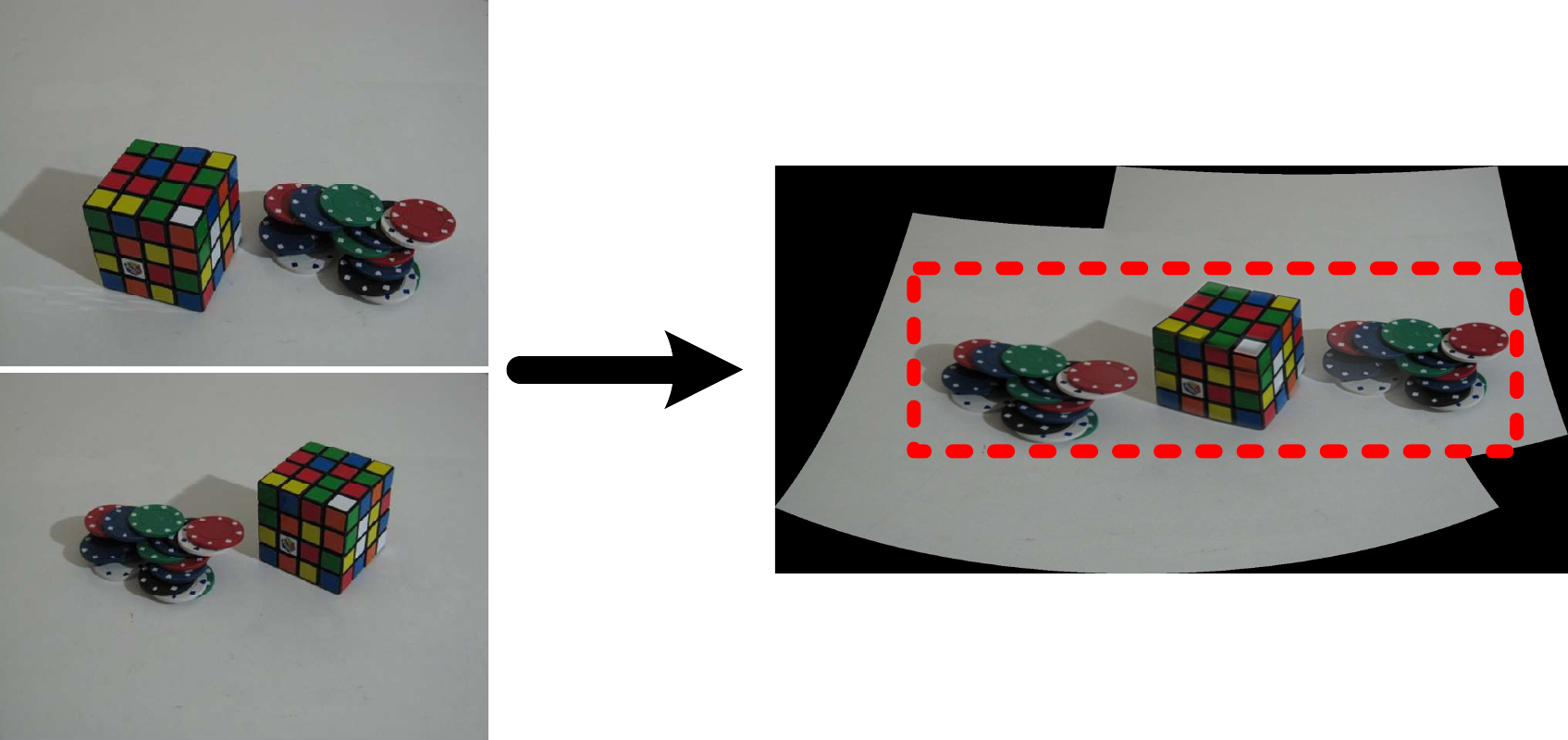}}
	\subfigure[Multiple local stitching]
	{ \includegraphics[width=0.9\linewidth]{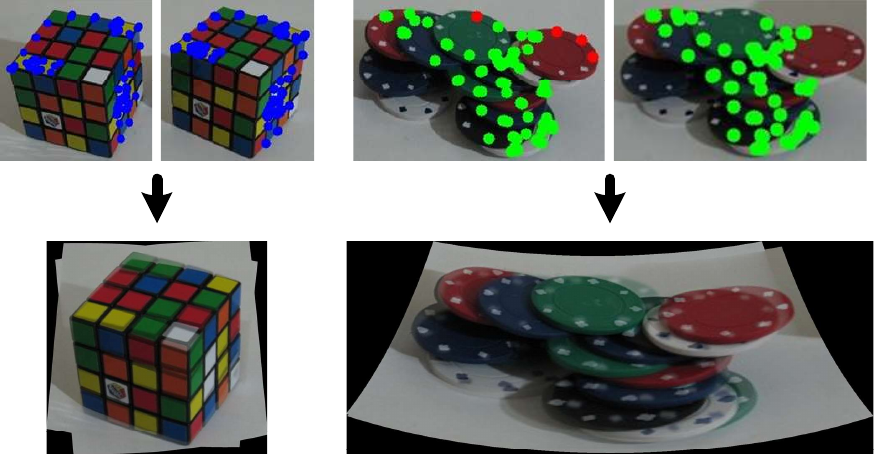}}
	\caption{Image stitching. The dot boxes in (a) illustrate some misalignment artifacts between two images where multiple structures , -i.e., to moving objects, are included. (b) shows the effectiveness of ic-GTM that recognizes multiple consistencies which are further leveraged to estimate some local transformations and stitch multiple objects separately.}
	\label{fig:stitch}
\end{figure}

The problem of image mosaicing (a.k.a. image stitching) has been extensively studied for static scenes where the alignment of two images is determined by a global transformation (homography matrix). However, traditional image mosaicing technique easily fails when applied to dynamic scenes as illustrated in Fig.~\ref{fig:stitch} (a). Misalignment or misregistration is inevitable when a single global transformation is insufficient to characterize the geometric relationship between the input pair.  The mosaicing result suffers from unnatural "ghosting" artifact (as highlighted by dot boxes). 

We propose to generalize the traditional image mosaicing problem into dynamic scenes. Such dynamic image mosaicing \cite{zhi2011toward} can support multi-frame image super-resolution and video mosaicing. Based on the developed multi-consistency matching or image alignment method, one can simply project multiple distinct objects in the source image by different local transformations. And accordingly, the mosaicing of stitching of each object can be guided by the corresponding local transformation as shown in Fig.~\ref{fig:stitch} (b). %Therefore, Gird-GTM is motivated to be integrated into some image stitching frameworks, which plays a role of recognizing the underlying consistencies located in separate structures and estimating corresponding transformations respectively.  

\subsection{Dynamic Object Tracking}
\begin{figure}[!t]
	\centering 
	\includegraphics[width=1\linewidth]{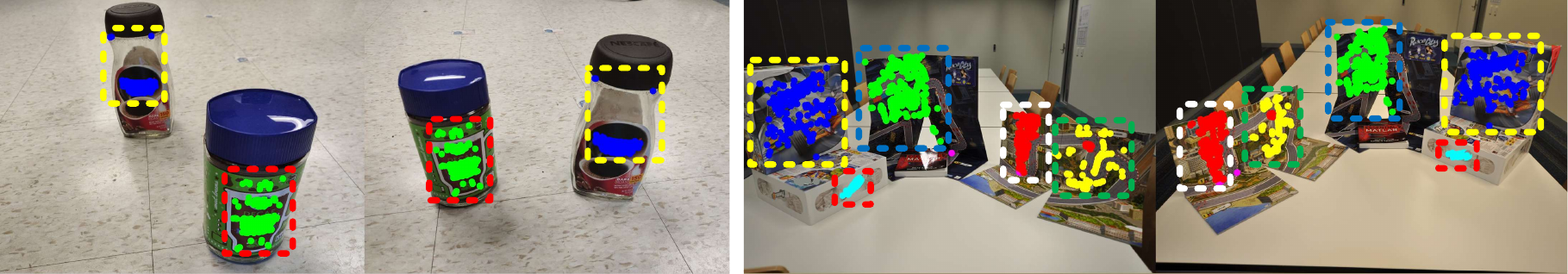}
	\caption{Generated proposals for dynamic tracking. The correspondences belonging to the same object pair are clustered and recognized by ic-GTM, which are represented by different colors. Some bounding box proposals are able to be generated, which provide a reliable prior knowledge for dynamic tracking. }
	\label{fig:co_seg}
\end{figure}

The other niche application of multi-consistency feature matching is dynamic object tracking in video. Although the problem of object tracking has also been widely studied, robust tracking of multiple objects has remained a long-open problem \cite{bernardin2008evaluating}. The accuracy of current state-of-the-art tracking algorithms is merely below 60\% \cite{kristan2017visual}. In some challenging scenario such as video captured from unmanned aerial vehicles (UAV), the task of multiple object tracking face several adversary factors -e.g., viewpoint changes, scale variations, and camera rotations. 

This work provides a new set of tools for tackling the problem of dynamic object tracking. As demonstrated in Fig.~\ref{fig:co_seg}, ic-GTM was capable of separately clustering the selected correspondences regardless of the large viewpoint changes. Robust feature correspondences as highlighted by different colors provide plausible bounding box proposals that can be used as the initial hypothesis by devoted object tracking algorithms. Due to space limitation, we will report more quantitative experimental results (e.g., VOT2017) in the future.

\section{Conclusion}
\label{sec:conclu}
In this paper, we presented an iterative clustering with Game-Theoretic Matching (ic-GTM) method focusing on selecting correct matches in context of multiple coherent correspondences. This method is robust to common nuisances and significantly outperforms other state-of-the-art approaches on both the multi-consistency feature matching task and single consistency feature matching task. In addition, to fill the gap of multi-consistency evaluation, we proposed a benchmark including a dataset set up in three scenes and three new metrics that are more reasonable for multi-consistency measurement. The code and benchmark will be available at: \url{https://github.com/sailor-z/ic-GTM}.

\section*{Acknowledgment}
This work was supported in part by the National Natural Science Foundation of China under Grant 61876211 and by the 111 Project on Computational Intelligence and Intelligent Control under Grant B18024.
%\begin{acknowledgements}
%If you'd like to thank anyone, place your comments here
%and remove the percent signs.
%\end{acknowledgements}

% Authors must disclose all relationships or interests that 
% could have direct or potential influence or impart bias on 
% the work: 
%
% \section*{Conflict of interest}
%
% The authors declare that they have no conflict of interest.

% BibTeX users please use one of
%\bibliographystyle{spbasic}      % basic style, author-year citations
\bibliographystyle{spmpsci}      % mathematics and physical sciences
\bibliography{mybibfile}   % name your BibTeX data base

% Non-BibTeX users please use
%\begin{thebibliography}{}
%
% and use \bibitem to create references. Consult the Instructions
% for authors for reference list style.
%
%\bibitem{RefJ}
% Format for Journal Reference
%Author, Article title, Journal, Volume, page numbers (year)
% Format for books
%\bibitem{RefB}
%Author, Book title, page numbers. Publisher, place (year)
% etc
%\end{thebibliography}
\end{sloppypar}
\end{document}